\documentclass[12pt,letterpaper]{article}
\usepackage{fullpage}
\usepackage[top=2cm, bottom=2cm, left=2.5cm, right=2.5cm]{geometry}
\usepackage{cite}
\usepackage{amsmath,amssymb,amsfonts,amsthm}
\usepackage{algorithmic}
\usepackage{graphicx}
\usepackage{textcomp}
\usepackage{commath}
\usepackage{hyperref}
\usepackage{algorithm}
\usepackage{algorithmic}
\usepackage{subfigure}
\usepackage{caption}
\usepackage{xcolor}
\usepackage{mathtools}
\usepackage[normalem]{ulem}
\usepackage{enumerate}

\newtheorem{assumption}{Assumption}
\newtheorem{corollary}{Corollary}
\newtheorem{lemma}{Lemma}
\newtheorem{definition}{Definition}
\newtheorem{remark}{Remark}
\newtheorem{theorem}{Theorem}
\usepackage{booktabs}

\newcommand\numberthis{\addtocounter{equation}{1}\tag{\theequation}}

\newcommand{\be}[0]{\begin{equation}}
	\newcommand{\ee}[0]{\end{equation}}
\newcommand{\ben}[0]{\begin{equation*}}
	\newcommand{\een}[0]{\end{equation*}}
\newcommand{\bena}[0]{\begin{eqnarray*}}
	\newcommand{\eena}[0]{\end{eqnarray*}}
\newcommand{\bea}[0]{\begin{eqnarray}}
	\newcommand{\eea}[0]{\end{eqnarray}}

\newcommand{\wt}{\widetilde{\theta}}

\newcommand{\tends}{\rightarrow}

\newcommand{\lm}[1]{\lim_{#1\tends\infty}}
\newcommand{\linfty}{{\cal L}_\infty}

\newcommand{\ov}[1]{\overline{#1}}
\newcommand{\RR}[0]{\mathbb R}
\newcommand{\N}{\mathcal{N}}
\newcommand{\R}{\mathbb{R}}
\newcommand{\tr}{\mathrm{Tr}}

\newcommand{\redc}[1]{\textcolor{black}{#1}}
\newcommand{\bluec}[1]{\textcolor{black}{#1}}
\newcommand{\brownc}[1]{\textcolor{black}{#1}}

\newcommand{\orangec}[1]{\textcolor{black}{#1}}

\newcommand{\remove}[1]{\textcolor{red}{\sout{#1}}}
\renewcommand{\remove}[1]{}
\renewcommand{\sout}[1]{}

\def\BibTeX{{\rm B\kern-.05em{\sc i\kern-.025em b}\kern-.08em
		T\kern-.1667em\lower.7ex\hbox{E}\kern-.125emX}}
\allowdisplaybreaks
\begin{document}
\title{Integration of Adaptive Control and Reinforcement Learning for Real-Time Control and Learning
\thanks{This work is supported by the Boeing Strategic University Initiative and the Air Force Research Laboratory, Collaborative
Research and Development for Innovative Aerospace Leadership
(CRDInAL), Thrust 3 - Control Automation and Mechanization grant FA
8650-16-C-2642.}
}
\author{Anuradha M. Annaswamy\thanks{AMA, AG, YC,  ST, and PF are with MIT, Cambridge, MA, 02139}, Anubhav Guha, Yingnan Cui, \\Sunbochen Tang, Peter A. Fisher, and Joseph E. Gaudio\thanks{JEG is with Aurora Flight Sciences, a Boeing Company, Cambridge MA 02140}
%
%
}
\date{}
\maketitle


\begin{abstract}
This paper considers the problem of real-time control and learning in dynamic systems subjected to parametric uncertainties. 
{We propose a combination of a Reinforcement Learning (RL) based policy in the outer loop suitably chosen to ensure stability and optimality for the nominal dynamics, together with Adaptive Control (AC) in the inner loop so that in real-time AC contracts the closed-loop dynamics towards a stable trajectory traced out by RL.} Two classes of nonlinear dynamic systems are considered, both of which are control-affine. The first class of dynamic systems utilizes equilibrium points 
and  a Lyapunov approach while second class of nonlinear systems uses contraction theory. AC-RL controllers are proposed for both classes of systems and shown to lead to online policies that guarantee stability using a high-order tuner and accommodate parametric uncertainties and magnitude limits on the input. In addition to establishing a stability guarantee with real-time control, the AC-RL controller is also shown to lead to parameter learning with persistent excitation for the first class of systems. Numerical validations of all algorithms are carried out using a quadrotor landing task on a moving platform. \remove{These results point out the clear advantage of the proposed integrative AC-RL approach.}
\end{abstract}

\section{Introduction}\label{s:intro}
This paper considers the problem of real-time control and learning of a class of dynamic systems with parametric uncertainties using a combination of Adaptive Control (AC) and Reinforcement Learning (RL).  \remove{The overall goals of  AC are to guarantee that the control input results in a closed-loop adaptive system that has globally bounded solutions, and to learn the true parameters \cite{Narendra05, Ioannou1996, Sastry_1989, Slotine1991, Krstic1995, Ast13,Ast07, Guay2003,hov2010}.} Over the past seven decades, {the field of AC has focused on ensuring} global boundedness with asymptotic tracking, parameter learning with persistent excitation, and robustness to nonparametric uncertainties such as disturbances and unmodeled dynamics \remove{have been ensured} for a large class of {linear and} nonlinear dynamic systems \cite{Narendra05,Ioannou1996,Sastry_1989,Slotine1991,Krstic1995,Ast13,Ast07,Guay2003,hov2010,Narendra_1986,Hussain_2017}. The field of RL has focused on the determination of a sequence of inputs that drives a dynamical system to minimize a suitable objective with minimal knowledge of the system model. The central structure of the RL solution revolves around a policy that is learned so as to maximize a desired reward \cite{Recht2019,Ber96,Watkins92,Sutton1998}.

Both AC and RL-based control methods have addressed the problem of control in the presence of uncertainty with a component of learning, but with entirely different approaches. AC methods have been proven to be effective in the 
enforcement of objectives for specific classes of problems, such as regulation and tracking, in real-time,  with analytical guarantees. In addition to accommodating parametric uncertainties, constraints on the control input magnitude \cite{karason1994,Lav04} and rate \cite{Gaudio_2019b} \bluec{can also be incorporated within an analytical framework.} They are however unable to directly guarantee the realization of long-term optimality-based objectives. RL- trained policies, on the other hand, can handle a broad range of objectives \cite{Sutton1998}, where the control policies are  learned \bluec{offline} in simulation, allowing for a near infinite number of agent-environment interactions to allow the policy to become near-optimal with guarantees of local stability \cite{Watkins92,jaakkola1993convergence,konda1999actor,lale2022reinforcement,jin2020provably,Berkenkamp2017}. In practice, however, offline policies trained in simulation often exhibit degenerate performance when used for real-time control due to modeling errors that can occur online \cite{Koos10,Tan18,Fulton18, Roy2017, Rajeswaran2017}. This paper takes a first step towards an integration of the AC and RL approaches so as to reduce this “sim-to-real” gap by realizing a combined set of advantages of both approaches.  

The AC-RL approach that we propose in this paper uses AC-based components in the inner-loop and RL-based components in the outer-loop. The AC-RL controller is designed such that the inner-loop AC accommodates the effect of parametric uncertainties in real-time and contracts the closed-loop dynamics towards a reference system through a nonlinear adaptive law. The RL {policy} in the outerloop, {which is used to determine the control input to the reference system,} is designed to lead to an almost-optimal control policy for the reference system, through training in simulation. {An important point to note is that this policy   is chosen so that local stability is ensured, which can be accomplished by introducing additional components in the underlying loss function (ex. \cite{Berkenkamp2017}).} 
Two classes of nonlinear dynamic systems are considered,  denoted as Problem 1 and Problem 2. The first class of dynamic systems utilizes equilibrium points and expansion forms around these points and employs a Lyapunov approach to establish convergence. The second class of nonlinear systems  uses contraction theory as the underlying framework \cite{lohmiller1998contraction}. In each class, an AC- RL controller is proposed for a class of parametric uncertainties, and the resulting closed-loop system is proved to be globally bounded with asymptotic properties of tracking and cost. In both problems, the states of the dynamic system are assumed to be accessible for measurement.

In the first class of systems, the RL component in the outer loop \bluec{ 
is chosen so as to lead to near-optimality of a desired cost function with analytical guarantees of }\bluec{local stability for a nonlinear reference system}\bluec{\cite{Berkenkamp2017}}. With this RL, and an AC-based design \bluec{that employs a high-order tuner (HT) in the inner-loop for parameter estimation}, we show that for a class of parametric uncertainties, the  closed-loop system, with the integrated AC-RL controller, can be guaranteed to be stable for all initial conditions of the true system. We also quantify the asymptotic behavior of its state, input, and overall cost. The advantages of the AC-RL controller are illustrated using numerical experiments of a quadrotor required to land on a moving platform using a nonlinear high-fidelity model, illustrating the reduction in the sim-to-real gap. Various extensions of the AC-RL approach are also proposed including a relaxation of the parametric uncertainties in the underlying nonlinearities, inclusion of multiple equilibrium points, non-affine systems, and incorporation of magnitude limits in the control input. In all cases except the last extension, the AC-RL ensures global boundedness;  in the last case, all initial conditions of the state and parameter estimate that lie inside a bounded domain are shown to lead to boundedness for an arbitrary nonlinear dynamic system.

In the second class of dynamic systems, denoted as Problem 2, we introduce  parametric uncertainties similar to the first class and employ contraction theory to establish convergence which utilizes tools in Riemannian geometry to analyze the convergence of differential dynamics by constructing differential Lyapunov functions \cite{lohmiller1998contraction,manchester2017control}. In contrast to prior work \cite{lopez2020adaptive}, we introduce multiplicative parametric uncertainties in the control input, \bluec{employ a HT for parameter estimation}, and an RL-based outer-loop {as before, with analytical guarantees of local stability} to synthesize the reference system. 

In both classes, the central component of the AC algorithm uses a HT which was first proposed by \cite{Morse_1992}, and later expanded in \cite{ortega1993,Evesque_2003} in an effort to develop stable low-order adaptive controllers. 
Independently, HT has also been sought after in the ML community, in an effort to obtain accelerated convergence of an underlying cost function and the associated accelerated learning of the minimizer of this function (see for example, \cite{Nesterov_1983,Nesterov_2018, Su_2016, Wibisono_2016, Zhang2018b}). The basic idea is to include momentum-based updates so as to get a faster convergence of the performance error. The HT algorithms in \cite{Morse_1992} have seen widespread applications in  machine learning  \cite{krizhevsky2017}. They were shown in \cite{Gaudio2021,Gaudio20AC,Mor2021} to lead to stability with time-varying regressors, and in \cite{Gaudio20AC} to have accelerated convergence with constant regressors in discrete-time. The first contribution of this paper is the demonstration of these properties of global boundedness and asymptotic behavior of the overall AC-RL controller in problems 1 and 2.

The second contribution of this paper is the demonstration of parameter learning in uncertain dynamic systems. We focus our attention primarily on the first class of dynamic systems with a scalar input. We show that under conditions of persistent excitation \cite{Narendra05}, the AC-RL controller with the HT guarantees that the parameter estimates converge to their true values, thereby allowing parameter learning. Since the HT inserts additional filtering actions, the proof of parameter convergence is a nontrivial result. Earlier results related to parameter convergence with HT required additional processing and time-scale transformations on the underlying regressors \cite{ortega1993}, which are avoided in this paper. {The main benefits of parameter learning are robustness \cite{Narendra_1986} and avoiding the bursting phenomenon \cite{Narendra05,annaswamy2023arcra}.} 

Efforts to combine AC and RL approaches have been highlighted in several recent works, which include \cite{Gaudio_2019a,Mat19, westenbroek2020adaptive, sun2021lyapunov} in discrete-time  
and \cite{Guha21,Richards2021, Dean_2018, fazel2018global} in continuous-time. Reference \cite{Richards2021} proposes the use of adaptive control for nonlinear systems in a data-driven manner, but requires offline trajectories from a target system and does not address accelerated learning or magnitude saturation. References \cite{Dean_2018, fazel2018global} study the linear-quadratic-regulator problem and its adaptive control variants from an optimization and machine learning perspective. In \cite{westenbroek2020adaptive} a reinforcement learning approach is used to determine an adaptive controller for an unknown system, while in \cite{sun2021lyapunov} principles from adaptive control and Lyapunov analysis are used to adjust and train a deep neural network. These approaches, including our earlier work in \cite{Guha21}, have not addressed a comprehensive treatment of an integrated AC-RL approach with analytical guarantees. 
While the use of an AC in the inner-loop and RL in the outer-loop is a common feature, all other elements in this paper including high-order tuner, magnitude saturation, multiple equilibrium points, non-affine systems, and contraction-theory based adaptive control design have been considered for the first time. None of these prior papers have addressed the problem of parameter convergence using the AC-RL controller, and this is also a key contribution of this paper. 

In summary, the contributions of the paper are the following, which are applicable to two classes of nonlinear dynamic systems 
with certain types of parametric uncertainties:
\begin{itemize}
    \item An AC-RL controller approach that ensures global stability with the AC in the inner loop employing a high-order tuner and RL component in the outer loop, with asymptotic quantification of its state, input, and overall cost with respect to a reference system. {The RL-based control policies are designed so that local stability is guaranteed and are near-optimal for the reference system.} 
    \item A guarantee of parameter learning in real-time using the AC-RL controller with persistent excitation. 
\end{itemize}

In Section \ref{sec:prob}, we describe the problem statements.  
Section \ref{sec:prelim} includes overviews of a range of underlying tools related to the RL and AC approaches. 
Sections \ref{sec:cont} and \ref{sec:problem_1_extensions} describe the AC-RL controller for Problem 1  and Section \ref{s:problem3} outlines the controller for Problem 2.  
Section \ref{sec:num_val} includes numerical {studies} of the AC-RL controllers for Problems 1 and 2. Proofs of Theorems~\ref{th:classic} and \ref{t:HAC-RL}-\ref{t:Contraction-HTAC} as well as all lemmas can be found in the Appendix. Proofs of Theorems 2 and 3 can be found in \cite{manchester2017control} and \cite{Morgan77a} respectively.

%
%

\section{Problem Statement}
\label{sec:prob}
We consider two classes of problems in this paper, for which we propose  control and learning solutions in real-time. In all problems, states are assumed to be accessible.

\subsection{Problem 1: Real-time control and learning for  systems with equilibrium points}
\label{sec:prob1}
Consider a continuous-time, deterministic nonlinear system described by the following dynamics:
\begin{equation}\label{eq:general_dynamics}
\dot{X}(t) = F(X(t), U(t))
\end{equation}
where $X(t)\in\RR^n$, and $U(t)\in\RR^m$. 
We define $x=X-X_0$ and $u=U-U_0$, where $(X_0,U_0)$ is an equilibrium point\footnote{In most of what follows, we suppress the argument $t$ for ease of exposition.}, i.e., 
$F(X_0,U_0)=0$. 
 Using a Taylor series expansion on \eqref{eq:general_dynamics} yields 
    \begin{equation}
        \dot{x} = Ax + Bu + f(x,u)\label{eq:general_dynamics2}
    \end{equation}
where $f: \mathbb{R}^n \times \mathbb{R}^m \to \mathbb{R}^n$, and 
$(A,B)$ is controllable. It should be noted that \eqref{eq:general_dynamics} can always be written in the form of \eqref{eq:general_dynamics2} for any analytic function $F$.  The following assumption is made about $f$:
    \begin{assumption}\label{a1} The higher order effects represented by the nonlinearity $f(x,u)$ in \eqref{eq:general_dynamics2} (a) lies in the span of $B$ and (b) are solely a function of the state $x$, i.e. the system in \eqref{eq:general_dynamics2} is control affine and $g(x)$ exists such that $Bg(x)=f(x,u)$: \end{assumption}
\be \dot x = Ax + B[u+g(x)]        \label{eq:affine}\ee
The goal in Problem 1 is the determination  of the control input $u(t)$ in real-time, when 
parametric uncertainties are present in the system dynamics in \eqref{eq:affine}, so as to minimize the cost function
\begin{equation}\label{eq:optimal_control}
\begin{aligned}
\min_{{\scriptscriptstyle u(t) \in {\cal U}, \; \forall t \in [t_0,t_0+T]}} \quad & \int_{t_0}^{t_0+T}c(x(t), u(t))dt\\
\end{aligned}
\end{equation}
subject to the dynamics in (\ref{eq:affine}), 
where ${\cal U}$ represents the set of all allowable inputs, and $c$ is a bounded \brownc{function that is piecewise continuous in $x$ and continuous in $u$}. 

\subsection{Problem 2: Real-time control for a class of nonlinear systems}
\label{sec:prob3}
Problem 2 corresponds to a class of control-affine nonlinear systems with parametric uncertainties which does not necessarily utilize the presence of an equilibrium point. This class is assumed to be of the form
\begin{equation}\label{equ:control_affine}
    \dot{X}(t) = f(X) + B(X)[U(t)]
\end{equation}
where $X(t) \in \mathbb{R}^n $ and $U(t) \in \mathbb{R}^m$. We note that \eqref{equ:control_affine} differs from the expansion form of \eqref{eq:general_dynamics2} and therefore allows both $f(X)$ and $B(X)$ to be nonlinear. The goal once again is to determine $U(t)$ in real-time which minimizes a cost function of $X(t)$ and $U(t)$, in the same form as \eqref{eq:optimal_control}, when parameter uncertainties are present in $f(X)$ and $B(X)$.

\section{Underlying Tools}\label{sec:prelim}

\subsection{An Offline Approach Based on Reinforcement Learning}\label{sec:prob-offline}
We define a reference system 
\begin{equation}\label{eq:reference-system-analog}
\dot{x}_r = Ax_r + Bu_r + f_r(x_r,u_r)
\end{equation}
where the subscript $r$ is used to denote signals and parameters of the reference system. We briefly describe the offline training procedure in an RL: 
First, it is assumed that the continuous-time dynamics in \eqref{eq:reference-system-analog} are sampled with sufficient accuracy, resulting in the discrete time dynamics
\begin{equation}\label{eq:tran}
\begin{aligned}
	x_{r, k+1} = h(x_{r, k}, u_{r, k})
\end{aligned}
\end{equation}
\noindent An appropriate numerical integration scheme ensures that this discrete-time formulation closely approximates the dynamics. The goal of RL is to learn a feedback policy $u_{r,k} = \pi(x_{r,k})$, with repeated training of $\pi(\cdot)$, so as to achieve the control objective of \eqref{eq:optimal_control} \cite{kaelbling1996reinforcement}.

In order to facilitate such a training, a simulation environment is assumed to be available so that each timestep, an observation $x_{r, k}$ is received, a control $u_{r,k}$ is chosen, and the resulting cost $c_k = c(x_{r, k}, u_{r, k})$ is computed. Repeating this process, a set of input-state-cost tuples $\mathcal{D} = [(x_{r, 1},u_{r, 1}, c_1), \ldots, (x_{r, N},u_{r, N}, c_N)]$ is formed. This data is used to train 
and update the policy $\pi$ \cite{kaelbling1996reinforcement}. Often the policy is  parameterized using neural networks so that $u_k = \pi_\theta (x_k)$, where $\theta$ denotes the weights of the neural network. 
A learning algorithm, often based on a gradient-descent approach, is used to adjust $\theta$ so that the expected accumulated cost is minimized \cite{Schulman2015}. In stochastic settings, the RL goal can be summarized as
\begin{equation}\label{eq:reward}
\begin{aligned}
	\min_{\theta} \quad & J(\theta)
	= \mathbb{E}_{\pi_\theta}\left[\sum_{k = 0}^T c_k \right ]
\end{aligned}
\end{equation}

\subsubsection{Analytical guarantees}
While majority of the RL literature is focused on extensive simulation studies, \bluec{a few papers have developed specific RL methods with analytical properties, which can be used for control of nonlinear systems (see for example, \cite{Watkins92,jaakkola1993convergence,konda1999actor,jin2020provably,lale2022reinforcement,Berkenkamp2017}). Reference \cite{Watkins92} shows convergence of Q-learning algorithms can be achieved using action-replay process, Ref. \cite{jaakkola1993convergence} develops convergence analysis based on stochastic approximation techniques that apply to both Q-learning and TD($\lambda$) algorithms, and Ref. \cite{konda1999actor} discusses local convergence results for actor-critic algorithms. Convergence analyses in \cite{Watkins92, jaakkola1993convergence, konda1999actor} are  established under a tabular setting where state and action spaces are both finite, which may be restrictive for control applications since most physical systems have continuous state space. Results in papers such as \cite{lale2022reinforcement,jin2020provably,Berkenkamp2017} avoid this restriction. 
Ref. \cite{lale2022reinforcement} proposes an RL algorithm provide stability guarantee for linear quadratic regulator (LQR) problems; 
results in \cite{jin2020provably} provide a proof of convergence of }\bluec{global optimality, and hence stable policy for linear MDPs and cost functions that are a linear combination of a set of known feature functions. In \cite{Berkenkamp2017}, RL is explored for general nonlinear systems, with the requirement that the  initial  policy is stable }\redc{ and a modification of the cost function that enforces Lyapunov stability based constraints. } 
\bluec{Overall, the approach in \cite{Berkenkamp2017} can be viewed as a policy design with a trade-off of optimality for closed-loop stability. We propose the use of such RL-based policies in the RL-part of our AC-RL controller.}

\subsection{Classical Adaptive Control Methods}
\label{sec:classic}
We start with an error model description of the form 
\be \dot e(t) = A_m e(t) + B\Lambda \left[\widetilde\Theta(t) \Phi(t)\right]\label{eq:errormodel}\ee
where $e(t)\in\RR^n$ is a performance error that is required to be brought to zero, such as tracking error, identification error, and state estimation error. $\widetilde\Theta(t)\in\RR^{m\times l}$ is a matrix of parameter errors that quantify the learning error. If the true parameter in the system dynamics is $\Theta^*\in\RR^{m\times l}$, and it is estimated as $\widehat\Theta(t)$ at time $t$, then $\widetilde\Theta=\widehat\Theta-\Theta^*$. $A_m$ is a Hurwitz matrix, $A_m$ and $B$ are known with $(A_m,B)$ controllable, and $\Lambda\in\RR^{m\times m}$ is an unknown parameter matrix that is positive definite. Finally $\Phi(t)\in\RR^l$ is a vector of regressors that correspond to all real-time information measured or computed from the system dynamics and controllers at time $t$. Such an error model is ubiquitous in adaptive control of a large class of nonlinear dynamic systems with parametric uncertainties \cite{Narendra05} including \eqref{eq:affine}. The following theorem summarizes the standard global stability result in AC:
\begin{theorem} \label{th:classic}
Let  $\Gamma\in\mathbb{R}^{m\times m}$ and $Q_a$ be symmetric positive-definite matrices with $P$ corresponding to the solution of the Lyapunov equation
\be A_m^TP+PA_m=-Q_a.\label{lyap}\ee
An adaptive law that adjusts the parameter estimate in real time as 
\be \dot{\widehat\Theta}= - \Gamma B^TPe \Phi^T \label{eq:gradadaptive}\ee
guarantees that $e(t)$ and $\widetilde\Theta(t)$ are bounded for any initial conditions $e(0)$ and $\widetilde\Theta(0)$. If in addition $\Phi(t)$ is bounded for all $t$, then $\lm{t}e(t)=0$.\end{theorem}


\subsection{High-order Tuners}
The core of the adaptive components proposed in this paper is based on  high-order tuners (HT) \cite{Morse_1992,ortega1993,Evesque_2003}. 
The idea behind HT is  summarized below.

Parameter identification in a linear regression problem of the form $y^*(t)={\theta^*}^T\phi(t)$ where $\theta^*\in\RR^n$ represents an unknown parameter, and $\phi(t)\in\RR^n$ is an underlying regressor that can be measured at each $t$ can be formulated as a minimization problem $L_t(\theta)=e^2(t)$ where $e=y - y^*$ and $y(t)=\theta^T\phi(t)$. A first-order tuner that can be used to solve the minimization problem is of the form $\dot{\theta}=-\gamma\nabla_{\theta} L_t(\theta)$. In \cite{Gaudio2021} \cite{Gaudio20AC}, a HT of the form
\begin{equation}\label{equ:HT}
\ddot{\theta} + \beta\dot{\theta} = -\frac{\gamma \beta}{\mathcal{N}_t}\nabla_{\theta} L_t(\theta)
\end{equation}
was proposed and shown to correspond to the Lagrangian
\begin{equation}\label{equ:HT-Lagrangian}
\mathcal{L}(\theta, \dot{\theta}, t) = e^{\beta (t-t_0)}\left( \frac{1}{2}\|\dot{\theta}\|^2 - \frac{\gamma \beta}{\mathcal{N}_t} L_t(\theta) \right).
\end{equation}
The benefits of the HT in \eqref{equ:HT} are that (a) it can be guaranteed to be stable even with time-varying regressors for a dynamic error model with a scalar control input \cite{Gaudio2021}, and (b) a particular discretization was shown in \cite{Gaudio20AC} to lead to an accelerated algorithm which reaches an $\epsilon$ sub-optimal point in $\widetilde{\mathcal{O}}(1/\sqrt{\epsilon})$ 
iterations for a linear regression-type convex loss function with constant regressors, as compared to the $\widetilde{\mathcal{O}}(1/\epsilon)$ guaranteed rate for the standard gradient descent algorithm. 
We employ elements of HT  in \cite{Gaudio2021} 
in the AC part of our AC-RL controller. 

\subsection{Contraction Theory}
Contraction theory provides tools to analyze the stability of nonlinear systems beyond attraction to an equilibrium point. Based on results in Riemannian geometry, control contraction metrics \cite{manchester2017control} have been developed to design a feedback controller that stabilizes a class of control-affine systems to a reference trajectory. A brief review of the salient features of this approach follows.

For an arbitrary pair of points $X_1, X_2 \in \mathbb{R}^n$, let $\zeta(X_1, X_2)$ denote the set of all smooth curves connecting $X_1$ and $X_2$, in which each path $c \in \zeta(X_1, X_2)$ is parameterized by $s \in [0, 1]$, i.e. $c(s): [0, 1] \to \mathbb{R}^n$, $c(0) = X_1, c(1) = X_2$. Let $\mathcal{S}_n^+$ be the set of $n\times n$ positive-definite matrices. A Riemannian metric $M(X): \mathbb{R}^n \to \mathcal{S}_n^+$ is a function that defines a smoothly varying inner product $<\cdot, \cdot>$ on the tangent space of $\mathbb{R}^n$.

A Riemannian metric $M(X)$ is uniformly bounded if there exists $\alpha_2 \geq \alpha_1 > 0$ such that $\alpha_1 I \leq M(X) \leq \alpha_2 I$. For a smooth path $c(s, t)$ between the pair $(X_1(t), X_2(t))$, which is a time-varying smooth curve parameterized by $s$ at any given $t\geq 0$, its Riemannian length and energy are defined as follows,
\begin{equation*}
L(c) \coloneqq \int_0^1 \|c_s\|_{c} ds, \quad E(c) \coloneqq \int_0^1 \|c_s\|^2_{c} ds
\end{equation*}
where $c_s \coloneqq \frac{\partial c(s, t)}{\partial s}$ and $\|c_s\|_{c} = \sqrt{c_s^T M(c(s, t)) c_s}$. A smooth curve is regular if $\frac{\partial c(s, t)}{\partial s} \neq 0$ for all $s\in [0, 1]$. By the Hopf-Rinow theorem, a smooth regular minimum-length curve (geodesic) $\gamma(s, t)$ exists and connects any pair $(X_1(t), X_2(t))$ \cite{manchester2017control}. Let the minimum Riemannian length of all paths connecting $X_1(t)$ and $X_2(t)$ be a Riemannian distance function, i.e. $d(X_1(t), X_2(t)) \coloneqq \inf_{c\in \Gamma(X_1(t), X_2(t))} L(c)$. The Riemannian energy of $\gamma(s, t)$ satisfies $d(X_1(t), X_2(t))^2 = E(\gamma) = L(\gamma)^2 \leq E(c)$ for all $c \in \Gamma(X_1(t), X_2(t))$, where the second equality follows from the fact that $\|\gamma_s\|$ is constant \cite{do2013riemannian}. 

We now consider the dynamic system in \eqref{equ:control_affine}, let $b_i$ be the $i$th row of $B(X)$ and $U_i$ be the $i$th element of control input vector $U(t)$. We can write the differential dynamics of \eqref{equ:control_affine} as 
\begin{equation}\label{equ:diff_sys}
\dot{\delta}_X = A(X, U) \delta_x + B(X) \delta_U, 
\end{equation}
where $A(X, U) = \frac{\partial f}{\partial X} + \sum_{i=1}^m \frac{\partial b_i}{\partial X}U_i$, and $\delta_x = \gamma_s$ is a differential element of the geodesic connecting a pair of points on two trajectories $X_1(t)$ and $X_2(t)$ in the state space $\mathbb{R}^n$. With these preliminaries established, we can now introduce the control contraction metric.
\begin{theorem}[\hspace{1px}\cite{manchester2017control}]
If there exists a uniformly bounded metric $M(X)$ and $\lambda > 0$ such that 
\begin{equation*}
\delta_X^T M B = 0 \implies \delta_X^T(A^T M + MA + \dot{M} + 2\lambda M)\delta_X \leq 0
\end{equation*}
holds for all $\delta_X \neq 0$, then system \eqref{equ:control_affine} is universally exponentially stabilizable with rate $\lambda$, and $M(X)$ is the control contraction metric. 
\end{theorem}
\begin{remark}
The above theorem states that under the metric $M(X)$, every tangent vector $\delta_X$ orthogonal to the span of $B(X)$ is naturally contracting with rate $\lambda$ \cite{manchester2017control}. Note that the differential dynamics in \eqref{equ:diff_sys} is linear time-varying, which reduces the complexity in finding a suitable $\delta_U$ that stablizes \eqref{equ:diff_sys} when $\delta_X^T M B\neq 0$. By computing path integrals of such a $\delta_U$, control contraction metrics $M(X)$ allows construction of stablizing controllers $U(t)$.
\end{remark}
\subsection{Persistent Excitation and Parameter Learning}\label{ss:def}
\begin{definition}[Persistent Excitation]\hspace{1sp}\cite{Narendra05}\label{def:PE}
A bounded function $\Phi:[t_0,\infty)\rightarrow\mathbb{R}^N$ is persistently exciting (PE) if there exists $T\hspace{-.045cm}>\hspace{-.045cm}0$ and $\alpha\hspace{-.045cm}>\hspace{-.045cm}0$ such that
\begin{equation}
\int_t^{t+T}\Phi(\tau)\Phi^T(\tau)d\tau\geq\alpha I,\quad \forall t\geq t_0.
\label{e:pe1}\end{equation}
\end{definition}
The definition of PE in \eqref{e:pe1} is equivalent to the following:
\be \frac{1}{T}\int_{t}^{t + T} \left|\Phi(\tau)^T w \right| d\tau \geq \epsilon_0, \forall\hbox{ unit vectors }w\in\RR^N, \forall t\geq t_0\label{e:pe2}\ee
for some $\epsilon_0 >0$. In what follows, we will utilize an alternate definition, which is equivalent to \eqref{e:pe1} and \eqref{e:pe2} if $\|\dot\Phi(t)\|$ is bounded for all $t$ \cite{Morgan_1977}:
\begin{definition}\label{d:pe}
$\Phi$ is PE if there exists an $\epsilon>0$  a $t_2$ and a sub-interval $[t_2,t_2+\delta_0]\subset[t,t+T]$ with
\be
\label{eq:pe}
\frac{1}{T}\left|\int_{t_2}^{t_2 + \delta_0} \Phi(\tau)^T w d\tau\right| \geq \epsilon_0, \forall\hbox{ unit vectors }w\in\RR^N, \forall t\geq t_0.\ee
\end{definition}

Adaptive control systems enable parameter learning by imposing properties of persistent excitation defined in Section \ref{ss:def}. We briefly summarize the classical result related to this topic, established in \cite{Morgan_1977,Morgan77a}.  The starting point is the same error model as in \eqref{eq:errormodel}, which contains two errors, $e(t)\in\RR^n$ is a performance error that can be measured, and $\wt^T(t)=\widetilde\Theta(t)\in\RR^{1\times l}$ is a parameter learning error. As before we assume that $A_m$ is known and Hurwitz, and $B$ is known. The following theorem summarizes this result: 
\begin{theorem}[\hspace{1px}\cite{Morgan77a}]\label{t:pe}  The solutions of the error dynamics in \eqref{eq:errormodel} together with the adaptive law in \eqref{eq:gradadaptive} lead to $\lm{t}\wt(t)=0$ if the regressor $\Phi$ satisfies the PE condition in Definition 2. \end{theorem}


\section{Integrated AC-RL solutions for real-time control and learning: Problem 1}
\label{sec:cont}

We address Problem 1 in this section, and propose an integrated AC-RL controller that combines adaptive control (AC) and reinforcement learning (RL) approaches.
\subsection{A Restatement of Problem 1}\label{s:acrl1}
We shall denote \eqref{eq:affine} when there are no parametric uncertainties as the reference system, and express it as 
   \begin{equation}
   \dot{x}_r = Ax_r + B[u_r + g(x_r)]\label{1b}
    \end{equation}
     We assume that $u_r$ is designed using RL:
    \begin{equation}
        u_{r} = \pi(x_r) \label{10}
    \end{equation}
    It should be noted that
    $\pi(x_r)$ in \eqref{10} does not necessarily cancel $g(x_r)$, but accommodates it so that the closed-loop system behaves in a satisfactory manner for the requisite control task. This is quantified in Assumption \ref{rl_policy}, a desirable property of RL controllers.
    
    \begin{assumption}\label{rl_policy} 
      RL is used to train a feedback policy $\pi: x \to u$ such that for a given positive constant $R_2$, a positive constant $R_1$ exists such that $||x_r(0)|| \leq R_1$ implies $||x_r(t)|| \leq R_2 \; \forall t$.
    \end{assumption}

\begin{remark}
    \bluec{Assumption 2 encapsulates the notion that, given a sufficiently long training period, a performant RL algorithm, and appropriate choices of hyperparameters, 
    the learned policy $\pi$ will be ``good" enough to maintain boundedness of the reference system states. As mentioned in Section \ref{sec:prob-offline}, for tabular settings where the action and state are finite \cite{Watkins92, jaakkola1993convergence}, and for linear dynamics and cost functions that are linear combinations of known feature functions \cite{jin2020provably}, the learned policy is optimal and therefore stable, which validates Assumption 2. For nonlinear dynamics, \cite{Berkenkamp2017} shows that a stable controller can be learned by optimizing a soft loss function that augments cost functions in \eqref{eq:optimal_control} with  Lyapunov stability based constraints. }\bluec{ Relaxing the soft loss constraints and expanding scope of RL for general nonlinear systems is one of the objectives in the area of {\em safe reinforcement-learning}, a topic of intense research activity. We note  that any of aforementioned frameworks including \cite{Berkenkamp2017} can be used to compute a stable RL policy $u_r$ in \eqref{10}.} \redc{In order to avoid the associated optimality gap induced by these soft loss constraints, methods such as PPO, shown to guarantee local optimality in \cite{holzleitner2021convergence}, 
    }\brownc{engineer the underlying cost function through the addition of high magnitude penalties only on out-of-set excursions.}
\end{remark}
     
\subsubsection{Parametric uncertainties}\label{ss:paramuncertain}
    We now introduce  two parametric uncertainties, one in the form of control effectiveness, i.e., $u$ gets perturbed as $\Lambda u$, and the second as a perturbation in $g(x)$. Loss of control effectiveness is ubiquitous in practical problems, due to unforeseen anomalies that may occur in real-time, such as accidents or aging in system components, especially in actuators. Parametric uncertainties in the nonlinearity $g(x)$ may be due to modeling errors. The following assumptions are made about these two uncertainties. 
\begin{assumption}\label{a2}   The nonlinearity $g(x)$ in \eqref{eq:affine} is parameterized linearly, i.e., $g(x)=\Theta_{n,r}\Phi_{n}(x)$ where $\Theta_{n, r}\in\mathbb{R}^{m \times l}$, and $\Phi_{n}(x)\in\RR^l$.
\end{assumption}
\begin{assumption}\label{a3} $\Lambda$ is symmetric and positive definite, with $\|\Lambda\|\leq 1$.
\end{assumption}
\color{black}
\begin{remark}
    Assumption \ref{a2} implies that $g(x)$ in \eqref{eq:affine} can be approximated accurately using $l$ basis functions. The use of basis functions to approximate a nonlinearity is commonplace in adaptive control: see e.g. \cite{Sanner_1992,Palanthandalam_2004,Lavretsky2013}. 
    For simplicity, we assume that the approximation is exact in Assumption~\ref{a2}, which will be relaxed in Section \ref{sec:relax}. Assumption \ref{a3} introduces a structure in the  loss of control effectiveness.
\end{remark}
  
  With Assumption \ref{a2} in place, a second parametric uncertainty is introduced where the nonlinear component $g(x)$ is perturbed as $\Theta_{n,r}\Phi_{n}(x)$ to $\Theta^{'}_{n}\Phi_{n}(x)$. With these parametric uncertainties, the  plant equation in \eqref{eq:affine} then becomes
    \begin{equation}
        \dot{x} = Ax + B\Lambda \left[u + \Lambda^{-1}\Theta^{'}_{n}\Phi_{n}(x)  \right]
        \label{1c}
    \end{equation}
    where
     $ B \in \mathbb{R}^{n\times m}, \Lambda \in \mathbb{R}^{m \times m}$, and $ \Theta^\prime_{n}\in \mathbb{R}^{m \times l}.
    $ 
        Altogether we note that Problem 1 is restated as the control of \eqref{1c} where $A$, $B$, and $\Phi_{n}(x)$ are known, but $\Lambda$ and $\Theta^\prime_{n}$ are unknown parameters.

\subsection{The AC-RL controller}\label{ss:acrl1}
  The goal of the AC-RL controller is to design a control input $u$ in \eqref{1c} such that the true system state $x$ converges to the reference system state $x_r$ in the presence of the aforementioned perturbations in $\Lambda$ and $\Theta^\prime_{n}$.  Subtracting \eqref{1b} from \eqref{1c}, we obtain the error dynamics
    
   \begin{equation}
       \dot{e} = A_He + B\Lambda [u - \Theta \Phi ]\label{52}
   \end{equation} 
   where 
   \bea
      e\coloneqq x-x_r, \Theta \coloneqq
            \Lambda^{-1}\left[I ,\;
       -\Theta^{\prime}_{n}\right], 
       \Phi \coloneqq \begin{bmatrix}  \Phi_r(u_r,x_r,x) \\*[.1in]
       \Phi_{n}(x)  \end{bmatrix} \label{53}\\
   \Phi_r(u_r,x_r,x) = u_r+ g(x_r) + \Theta_{l,r}e \nonumber\eea
and $\Theta_{l, r}$ is such that $A_H \coloneqq A + B\Theta_{l,r}$ is Hurwitz. In \eqref{52}, $\Theta\in\mathbb{R}^{m\times (m+l)}$ corresponds to an unknown parameter matrix and $\Phi(t)\in\mathbb{R}^{m+l}$ is the regressor vector used for adaptation. The error equation \eqref{52} is central to the development of the AC-RL algorithms derived in the following subsections.
We now propose the AC-RL controller.
\begin{align}
    u &= \widehat\Theta(t)\Phi(t)
     \label{eq:adaptive_control}\\
      \dot{\Xi}&=-\gamma B^T Pe\Phi^T
      \label{eq:adaptive_param1}\\
    \dot{\widehat\Theta}&=-\beta(\widehat\Theta-\Xi)\N_t,\label{eq:adaptive_param3}
\end{align}
where \begin{align}
  \N_t &= 1 + \mu\Phi^T\Phi
  \label{eq:4}\\
    \mu&\geq \frac{2\gamma}{\beta}\|PB\|_F^2{\|\Omega\|_2^2}
    \label{eq:mu}
\end{align}
{ $\Omega$ is such that $\Omega^T\Omega = \Lambda$} 
, $\|\cdot\|_F$ in \eqref{eq:mu} denotes the Frobenius matrix norm, 
 and $P = P^T\in\RR^{n \times n}$ is a positive definite matrix that solves the equation $A_H^T P + P A_H = -Q_H$, where $Q_H$ is a positive-definite matrix and $Q_H\geq 2I$. 
 The following theorem is the first main result of this paper:

\begin{theorem} \label{t:HAC-RL}
Under Assumptions \ref{a1}-\ref{a3}, the closed-loop adaptive system specified by the plant in \eqref{1c}, the reference system in \eqref{1b},  and the adaptive controller in \eqref{eq:adaptive_control}-\eqref{eq:adaptive_param3} leads to  bounded solutions for any initial conditions $x(t_0)$, with $\lm{t}e(t)=0$.
\end{theorem}

\bluec{Using the result in Theorem \ref{t:HAC-RL}, we quantify the asymptotic property of AC-RL controller in Corollary \ref{cor:cost} where $o(\cdot)$ is defined as in \cite{hardy1979introduction}. For this purpose, we define the ideal control input as $u^*\coloneqq \Theta \Phi$. From \eqref{52}, it is easy to see that if $u=u^*$, then $x(t)\tends x_r(t)$ as $t\rightarrow\infty$.} 
\begin{corollary} \label{cor:cost} \bluec{If $\dot\Phi$ is bounded, the AC-RL controller ensures that (i) $\lm{t}|u(t)-u^*(t)|=0$ and
\be \mathrm{(ii)} \; \int_{t_0}^{t_0+T} \left|c(x(t),u(t))-c(x_r(t),u^*(t))\right|dt=o(T) \label{adaptive-cost}\ee}
\end{corollary}

We make several important observations about the AC-RL controller: 
\begin{enumerate}\itemsep 0pt
\item The parameter update in (\ref{eq:adaptive_param1}) - (\ref{eq:adaptive_param3}) is a second-order tuner,  and an extension of our earlier results in \cite{Gaudio2021} to the multivariable case. Two different regressors are employed in the adaptive control input, $\Phi_r$ in \eqref{eq:adaptive_control} and $\Phi_{n}$, which are utilized to address the two different sources of parametric uncertainties $\Lambda$ and $\Theta_{n}$. The first regressor component, $\Phi_r$, comes predominantly from the RL-component, where 
$g(x_r)$ can be computed as $(B^T B)^{-1} B^T(\dot{x}_r - A x_r) - u_r$ using \bluec{measurements from the reference system and numerical approximations.} \redc{Alternately, $g(x_r)$ can be computed using comparisons of simulations of the reference system for infinitesimal and large inputs. It should be noted that these computations are offline, and therefore do not add to the computational burden in the AC-RL.} The second regressor accommodates the uncertainty $\Theta^\prime_{n}$ in $g(x)$ (and employs assumption \ref{a2}). 
 The additional feedback from $e$ in \eqref{53} is essential in guaranteeing global stability.

\item {The RL-policy is required to satisfy Assumption 2, such as the one in \cite{Berkenkamp2017}.}
\item In comparison to a purely AC approach, in the AC-RL controller the RL plays the role of a reference model that is nonlinear with a controller that is nonlinear as well and trained so as to elicit desirable properties from the reference system. \redc{It should also be noted that a purely AC approach is agnostic to $u_r$, relegates the design of $u_r$ to an outer-loop, and essentially focuses only on the innerloop stability and tracking of $x_r(t)$.}

\item If there are no parametric uncertainties, and if the initial conditions of (\ref{1c}) are identical to those of (\ref{1b}), then the choices of $\widehat{\Theta}(0)= [I, \Theta_{n,r}^T]^T$ and $\Xi(0)=0$ ensures that the AC-RL control $u(t)\rightarrow u^*(t)\rightarrow u_r(t)$, 
thereby accomplishing \eqref{eq:optimal_control}. That is, the AC-RL policy coincides with the near-optimal RL-based policy. When there are parametric uncertainties, the AC-RL policy is \redc{necessarily} modified; $u(t)\tends u^*(t)$ but $u^*(t)\not\rightarrow u_{r}(t)$. 
Theorem \ref{t:HAC-RL} guarantees that with this modified policy, the closed-loop system state $x$ is globally bounded, sans any online training, and converges to $x_r$. Corollary \ref{cor:cost} precisely quantifies the corresponding cost of the AC-RL controller. 

\end{enumerate}
\begin{remark} Alternate methods rather than RL for determining the feedback policy may be used as well provided they satisfy Assumption \ref{rl_policy}. Methods based on MPC may be used as well, but may require additional structures on the cost function in the form of terminal costs \cite{kouvaritakis2016model}. The same comment is applicable for {other classical nonlinear control methods}{, but these} may involve additional assumptions or restrictions. Furthermore, MPC generally requires online computation, which might be challenging especially when system dynamics is nonlinear, while RL can be trained offline and deployed online with reduced computational burden. \end{remark}

\subsection{A comparison of AC-RL and RL-based controllers}
\sout{The problem we have considered in this section is the control of nonlinear dynamic systems with a single equilibrium point, which can always be written in the form of \eqref{eq:general_dynamics2}. The AC-RL controller we have proposed can guarantee that for any initial conditions $x(t_0)$ and any parametric uncertainties that satisfy Assumptions \ref{a1}, \ref{a2}, and \ref{a3}, the overall closed-loop system will be globally bounded, and that $x(t)$ will track the state of a reference system \eqref{1b} that satisfies Assumption \ref{rl_policy}.} We make a few comments regarding {Assumptions \ref{a1}-\ref{a3}} \sout{these assumptions} and the global properties of this AC-RL controller below.


\subsubsection{Global properties of the AC-RL controller vs. local properties of the RL-controller} The focus is on nonlinear dynamic systems with parametric uncertainties that satisfy Assumption \ref{a1}, \ref{a2}, and \ref{a3} and an RL-component that satisfies \ref{rl_policy}. Suppose we start with $x(t_0)$ with $||x(t_0)||\leq R_1.$ Suppose that the parametric uncertainties are such that $x(t)$ departs from ${\cal X}$ for some $t>t_0$. In such a case, an RL controller alone will not suffice. 
In contrast, the AC-RL controller will accommodate these departures, modify the policy from $u_r(t)$ to $u^*(t)$, and ensure that $$||x(t)||\leq (1/\lambda(P_{{\rm min}}))V_0+R_1\qquad \forall t\geq t_0$$ where $V_0=V((e(t_0),\Xi(t_0)-\Theta,\Theta(t_0)-\Theta)$, where $V$ is defined as in \eqref{eq:lyap}. This accommodation was possible because of the change in the control input from $u_r$ to $u$ in \eqref{eq:adaptive_control}. It should be noted that the AC-RL controller relaxes the restriction of the RL-approach that the state always remains in a compact set, allow parametric uncertainties to be introduced at any $t_0$, provided they are in the form specified in Assumptions \ref{a1}, \ref{a2}, and \ref{a3}, and provided the RL approach satisfies Assumption \ref{rl_policy}. As this is a nontrivial class of nonlinear problems that can occur in practice, this extension from a local result with RL-approaches to a global result with AC-RL approach proposed is an important contribution, and bridges the sim-real gap for this class of nonlinear problems. The benefit of using RL rather than other control approaches is that we can address tasks that correspond to arbitrary cost functions.
\bluec{The cost of using RL is the requirement that Assumption \ref{rl_policy}
 is satisfied.}
 
\subsubsection{Assumptions 1 to 4}

Assumption \ref{rl_policy} was needed for the RL part of the controller, while the other three assumptions were needed for the AC part of the controller. As the adaptive controller is designed to function in real time, certain structures needed to be imposed on the type of uncertainties that the AC can deal with, which led to Assumption \ref{a1} (that required uncertainties to lie in the span of the control matrix $B$), Assumption \ref{a2} (that required the nonlinearities to be represented using a linear network with a bounded approximation error), and Assumption \ref{a3} (which required certain analytic properties in the linear parametric uncertainty $\Lambda$). 
\subsubsection{Optimality of the control policy $u^*$}
While we have established the asymptotic nature of $u$ in the form of $u^*$, we note that $u^*(t) \neq u_{opt}(t)$, where
$$u_{opt}(t) = argmin_{u(t) \in {\cal U}, \; \forall t \in [t_0,t_0+T]} \int_{t_0}^{t_0+T}c(x(t), u(t))dt$$
for the plant \eqref{1c}.
\bluec{The reason for this difference can be argued thus: 
The input $u^*$ is designed to make the closed-loop system stable with uncertainties with the reference system as the target response. But such a $u^*$ is optimal with respect to the  cost function in \eqref{eq:optimal_control} subject to the reference system in \eqref{1b} (i.e. no uncertainties), but not optimal with respect to the perturbed system in \eqref{1c} (with uncertainties). While choosing the reference system in \eqref{1b} as a target response provides tractability in the form of global stability, it may not be the right choice for the compromised true system in \eqref{1c} from the point of view of optimality. A tractable design of a reference system that assures both stability (using AC methods) and optimality (using RL methods) is a topic for future research.
}

\subsection{Learning in AC-RL controllers with persistent excitation}
Section \ref{ss:acrl1} focused on the control solution, and showed global boundedness with a cost as in \eqref{adaptive-cost}. In particular we showed in corollary \ref{cor:cost} that $\widetilde \Theta(t)\Phi(t) \rightarrow 0$ asymptotically, where $\widetilde\Theta=\widehat\Theta-\Theta$. 
As the vector $\Phi$ can be arbitrary, $\widetilde\Theta(t)$ can remain orthogonal to $\Phi(t)$, and need not go to zero asymptotically, i.e., there is no guarantee that parameter learning will take place.
In this section, we establish  learning with the AC-RL algorithm by imposing additional conditions of persistent excitation on $\Phi$. We limit our discussion to the case when $u(t)$ is a scalar. 

The starting point is the error equation in \eqref{52} and the AC-RL controller in \eqref{eq:adaptive_control}-\eqref{eq:mu}.  
With a scalar input, $m=1$, we obtain that $\Lambda\in\mathbb{R}^+$ (using Assumption \ref{a3}), $\Theta\in\mathbb{R}^{1\times (l+1)}$, $\Phi_r(t)\in\mathbb{R}$, and $\Phi_{n}(t)\in\mathbb{R}^l$ in \eqref{53}. 
Defining $\tilde\theta=(\widehat\Theta-\Theta)^T$, $\tilde\vartheta=(\Xi-\Theta)^T$ and $B_0 = B\Lambda$, we rewrite \eqref{52}, \eqref{eq:adaptive_param1} and \eqref{eq:adaptive_param3} as
\begin{equation}
  \label{eq:34}
  \dot e = A_H e + B_0\tilde{\theta}^T\Phi
\end{equation} and
\begin{align}
  \label{eq:adp1}
  \dot{\tilde{\vartheta}} &= -\gamma\Phi e^T PB\\
  \dot{\tilde{\theta}} &= -\beta(\tilde{\theta} - \tilde{\vartheta})\N_t\label{eq:adp2}
\end{align}
where $\N_t = 1 + \mu\Phi^T \Phi$, $\mu \geq 2\gamma\|PB\|^2/\beta$. As before, the matrix $P$ solves $A_H^T P + PA_H = -Q$ and $Q \geq 2I$ is a symmetric, positive-definite matrix. The constants $\gamma$ and $\beta$ are positive. Let $x_1(t) = [e(t)^T, (\tilde{\theta}(t) - \tilde{\vartheta}(t))^T]^T$ and $z(t) = [x_1(t)^T, \tilde{\vartheta}(t)^T]^T$. 

We note that the  stability result related to the AC-RL controller stated in Theorem 4 guarantees that $z(t)$ is uniformly bounded, which follows from a Lyapunov function of the form
  \begin{equation}
    \label{eq:lyap3}
    \begin{split}
      V &= \frac{\Lambda}{\gamma}\tilde\vartheta^T  \tilde\vartheta+\frac{\Lambda}{\gamma}\left[(\tilde\vartheta-\tilde\theta)^T  (\tilde\vartheta-\tilde\theta)\right] + e^T Pe
    \end{split}
  \end{equation}
  whose derivative is bounded by
  \be\label{e:dotV}
    \dot{V} \leq -\frac{2\beta\Lambda}{\gamma}\|\tilde\theta - \tilde\vartheta\|^2 - \Lambda\|e\|^2 
  \ee
  Noting that our goal is to show that $\lm{t}\tilde\theta(t)=0$, it suffices to show that $V(t)\rightarrow 0$ as $t\tends\infty$. As the time-derivative $\dot V$ in \eqref{e:dotV} is only negative semi-definite, additional conditions of  persistent excitation are utilized to show parameter learning, and corresponds to the second contribution of this paper. 
  
The following lemmas are useful in proving the second contribution of this paper, which is stated in Theorem \ref{theo:1}. We note that Theorem \ref{t:HAC-RL} guarantees that $z(t)$ and $\Phi(x,e,u_r)$ are bounded, making the properties of persistent excitation applicable for what follows.
\begin{lemma}
  \label{lemma:1}
  Let $\epsilon_1 > \epsilon_2 > 0$, then there is an $n = n(\epsilon_1, \epsilon_2)$ such that if $z(t) = \left[x_1(t)^T, \tilde{\vartheta}(t)^T\right]^T$ is a solution with $\|z(t_1)\| \leq \epsilon_1$ and $S = \left\{t\in[t_1, \infty)|\|x_1(t)\| > \epsilon_2\right\}$, then $\mu(S)\leq n$ where $\mu$ denotes Lebesgue measure.
\end{lemma}

\begin{lemma}
  \label{lemma:2}
  Let $\delta > 0$ and $\epsilon_1 > 0$ be given. Then there exist positive numbers $\epsilon$ and $T$ such that if $z(t)$ is a solution with $\|z(t_1)\|\leq \epsilon_1$ and if $\left\|\tilde{\vartheta}(t)\right\|\geq\delta$ for $t\in[t_1, t_1 + T]$, then there exists a $t_2\in[t_1, t_1 + T]$ such that $\|x_1(t_2)\|\geq \epsilon$.
\end{lemma}

\begin{lemma}
  \label{lemma:3}
  Let $\epsilon_1$ and $\delta$ be given positive numbers. Then there is a $T = T(\epsilon_1, \delta)$ such that if $z(t)$ is a solution and $\|z(t_1)\|\leq \epsilon_1$, then there exist some $t_2\in[t_1, t_1 + T]$ such that $\left\|\tilde{\vartheta}(t_2)\right\|\leq \delta$.
\end{lemma}

\begin{theorem}
  \label{theo:1}
  If $\Phi(t)$ satisfies the persistent excitation property in \eqref{eq:pe}, then the origin $(e=0,\tilde\vartheta=0,\tilde\theta=0)$ in \eqref{eq:34}-\eqref{eq:adp2} is uniformly asymptotically stable.
\end{theorem}

The  proof of Theorem \ref{theo:1} stems from the three lemmas listed above, which represent the three main steps. Lemmas \ref{lemma:1} and \ref{lemma:2} establish that $x_1(t)$ cannot remain small over the entire period of persistent excitation. Lemma \ref{lemma:3} then leverages this fact to show that this leads to the parameter error $\tilde\vartheta(t)$ to decrease. Together this allows the conclusion of u.a.s. of the origin \eqref{eq:34}-\eqref{eq:adp2} and therefore that $\lm{t}\tilde\theta(t)=0$. As is apparent from the details of the proof provided in the Appendix, first principles-based arguments had to be employed in order to derive this result. No standard observability properties or time-scale transformations as in \cite{ortega1993} have been employed; these are inadequate as the error model structure in \eqref{eq:34} includes system dynamics and no filtering techniques are used to convert the error model to a static linear regression model. Extensions to multivariable inputs are a topic for future work{. The results of this section also indicate that the RL-based policy $u_r$ should be such 
that various components of $\Phi(t)$ satisfy the PE condition.} \remove{and may need persistent excitation concepts to matrix regressors.}

\section{Extensions of the AC-RL controller for Problem 1} \label{sec:problem_1_extensions}
\subsection{Relaxation of Assumption 3}\label{sec:relax}
\begin{assumption}\label{a22}   The higher order term in \eqref{eq:affine} is parameterized linearly, i.e., $g(x) = \Theta_{n,r}\Phi_{n}(x) + d(t)$ where $\Theta_{n, r}\in\mathbb{R}^{m \times l}$, $\Phi_{n}(x)\in\RR^l$, and $d(t)$ denotes an approximation error with $\|d(t)\| \leq d_{max}$ for all $t$. \end{assumption}
\begin{remark}
   \bluec{While Assumption \ref{a22} is a relaxation of Assumption \ref{a2}, the more general case of interest is when $d$ to be dependent on the state $x$, i.e. not known to be bounded {\em a priori}. This extension is relegated to future work.}
\end{remark}

 We note that with Assumption \ref{a2} replaced by Assumption \ref{a22}, Eq. \eqref{1c} is modified as \begin{equation}
        \dot{x} = Ax + B\Lambda \left[u + \Lambda^{-1}\Theta^{'}_{n}\Phi_{n}(x) + \Lambda^{-1}d \right]
        \label{1c2}
    \end{equation}
    with the resulting error equation modified from \eqref{52} as 
 \begin{equation}
       \dot{e} = A_He + B\Lambda [u - \Theta \Phi + \overline{d}]\label{522}
   \end{equation} 
   where $\overline d=\Lambda^{-1}d$ with $|\overline d|\leq \bar d_{max}$.

The AC-RL controller in this case remains as in \eqref{eq:adaptive_control} and \eqref{eq:adaptive_param3}, but the tuner in \eqref{eq:adaptive_param1} is modified as 
\begin{align}
      \dot{\Xi}&=-\gamma B^T Pe\Phi^T-f_1(\Xi,
      \widehat\Theta)\label{eq:adaptive_param12}\\
 f_1&= \left(2\Xi-\widehat\Theta\right) \label{eq:mod}
\end{align}
with all other quantities defined as in Section \ref{sec:cont}. The analytical guarantee for this AC-RL controller is summarized in the following theorem:
\begin{theorem} \label{t:HAC-RL2}
Under Assumptions \ref{a1}, \ref{rl_policy}, \ref{a3}, and \ref{a22}, the closed-loop adaptive system specified by the plant in \eqref{1c}, the reference system in \eqref{1b},  and the adaptive controller in \eqref{eq:adaptive_control}-\eqref{eq:mu}, and \eqref{eq:adaptive_param1} replaced by \eqref{eq:adaptive_param12}-\eqref{eq:mod}, leads to globally bounded solutions, with $e(t)=O(D)$, where $D=\max(\bar d_{max},\|\Theta\|)$.
\end{theorem}

\subsection{MSAC-RL Controller}\label{sec:MSAC-RL Controller}
As control inputs are often subject to magnitude saturation, we propose another extension, Magnitude Saturated Adaptive Controller (MSAC)-RL, which builds on the ideas introduced in \cite{karason1994}. The saturated control input into the true plant is calculated as:
\be u_i(t)=u_{i, \max} {\rm sat}\left(\frac{u_{i, c}(t)}{u_{i, \max}}\right)\label{eq:sat}
\ee
where $u_c(t)$ denotes the output of the controller and $u_{i,\max}$ is the allowable magnitude limit on $u_i$, chosen as \bluec{$\|{\cal U}\|$ 
where ${\cal U}$ is the set of all allowable inputs}. This induces a saturation-triggered disturbance $\Delta u$ vector defined by
\be \Delta u(t)=u_c(t)-u(t)\label{eq:sat-dist}\ee
It is easy to see that $\Delta u(t)=0$ when the desired control $u_c(t)$ does not saturate. The output of the controller, $u_c$, is given by:
\begin{equation}\label{eq:uccontrol}
    u_c = \widehat\Theta(t)\Phi(t)    
\end{equation}
The presence of the disturbance $\Delta u$ causes the error equation to vary from \eqref{52} to 
\begin{equation}
       \dot{e} = A_He + B\Lambda [u_c-\Delta u - \Theta \Phi]\label{52sat}
   \end{equation}
We introduce a new performance error $e_a$ in order to accommodate the disturbance $\Delta u$ as follows:
\be \dot e_a =A_He_a +BK_a(t)\Delta u \label{eq:new-error-aux}\ee
where $K_a$ is an estimate for $\Lambda$ which is adjusted according to \eqref{eq:msac_eqs}. The error $e_a$ leads to a new augmented error $e_u=e-e_a$:
\begin{equation}
\begin{aligned}\label{eq:new-error}
    \dot e_u &= A_H e_u + B\Lambda\widetilde\Theta(t)\Phi(t)
    - B(\Lambda + K_a)\Delta u
\end{aligned}
\end{equation}
This suggests a different set of adaptive laws, 
\bea 
\dot{\Xi}&=&-\gamma B^T Pe_u\ov\Phi^T \label{eq:msac_eqs}\\
    \dot{\ov\Theta}&=&-\beta(\ov\Theta-\Xi)\N_t, \label{eq:msac_eqs_2}\eea
where $\ov\Theta=[\widehat\Theta, -K_a]$ and $\ov \Phi = [\Phi^T, \Delta u^T]^T$, $\gamma$ and $\beta$ are positive constants, and $\N_t$ defined as in \eqref{eq:4} with $\Phi$ replaced by $\ov\Phi$ and  $\mu$ as in  \eqref{eq:mu}. The following theorem provides the analytical guarantees for this MSAC-RL controller.
    
\begin{theorem}\label{t:MSAC-RL}
    Under Assumptions \ref{a1}-\ref{a3}, the closed-loop adaptive system specified by the plant in \eqref{1c}, the reference system in \eqref{1b},  the magnitude constraint in \eqref{eq:sat} and the MSAC-RL controller given by \eqref{eq:uccontrol}, (\ref{eq:msac_eqs}) and \eqref{eq:msac_eqs_2} leads to 
    \begin{enumerate}
    	\item[(i)] globally bounded solutions, with $\lim_{t \to \infty} \|e(t)\| = 0$ if the target system in \eqref{1c} is open-loop stable.
    	\item[(ii)] bounded solutions for all initial conditions $x(0)$, $\Xi(0)$, $\widehat\Theta(0)$ and $K_a(0)$ in a bounded domain, with $\|e(t)\| = O[\int_0^t\|\Delta u(\tau)\|d\tau]$ if the target system in \eqref{1b} is not open-loop stable.
	\end{enumerate}
\end{theorem}
\begin{remark} As in Section \ref{sec:relax}, if we relax Assumption \ref{a2} in the form of Assumption \ref{a22}, then the asymptotic result in Theorem \ref{t:MSAC-RL}(i) is replaced by convergence to a compact set as in Theorem \ref{t:HAC-RL2}. It should be noted that in both Section \ref{sec:relax} and \ref{sec:MSAC-RL Controller}, the proposed controller still remains global.\end{remark}


\subsection{Multiple Equilibrium Points}
Suppose the system in \eqref{eq:general_dynamics} has $p$ equilibrum points $(X_1, U_1), \dots, (X_p, U_p)$, so that $F(X_i, U_i) = 0$ for $i = 1, \dots, p$. Define $x_i = x - X_i$, $u_i = u - U_i$. Denoting the state and action spaces as ${\cal X}$ and ${\cal U}$, 
respectively, we partition the composite domain $\mathbb{D} = {\cal X} \times {\cal U}$ 
into $p$ disjoint subsets $S_1, \dots, S_p$, such that:
    \begin{align*}
        &\mathbb{D} = S_1 \cup S_2 \cup \dots S_p\\
        &S_i \cap S_j = \emptyset, i \neq j
    \end{align*}
One can then express \eqref{eq:general_dynamics} in the presence of parametric uncertainties as
\begin{equation}
    \dot{x} = \begin{cases} 
  A_1x_1 + B_1[\Lambda u_1+g(x_1)], & x_1, u_1 \in S_1 \\
  &\vdots \\
  A_px_p + B_p[\Lambda u_p+g(x_p)], & x_p, u_p \in S_p.
\end{cases}
\end{equation}
If we now consider the effect of parametric uncertainties in $u_i$ and $g(x_i)$ as in Section~\ref{sec:prob-offline}, it is easy to have a switching set of controllers as in \eqref{eq:adaptive_control}-\eqref{eq:adaptive_param3} that are invoked when the trajectories enter the set $S_i$. A corresponding stability result to Theorem~\ref{t:HAC-RL} can be derived,  provided the command signal is such that the dwell time in each set $S_i$ exceeds a certain threshold, which is not discussed here. We summarize the overall AC-RL controller in Algorithm \ref{alg:me_1}, which specifies the discrete time implementation of the overall AC-RL Controller. $\Delta t$ denotes the integration timestep and the AdaptiveControl function corresponds to the adaptive control input; in the single equilibrium case, this function corresponds to equations \eqref{eq:adaptive_control}-\eqref{eq:mu}. $F$ and $F_r$ represent the velocity vector fields of the target and reference dynamical systems, respectively. That is, $F$ and $F_r$ are the right hand sides of equations \eqref{eq:general_dynamics2} and \eqref{eq:reference-system-analog}, respectively.
\begin{algorithm}
\caption{Multiple Equilibrum AC-RL}\label{alg:me_1}
\begin{algorithmic}
\REQUIRE $F, F_r, \pi, x_r, x$
\WHILE {running}
\STATE $u_r = \pi(x_r)$
\STATE $x_i \leftarrow x - X_i$
\STATE $u_i \leftarrow u - U_i$
\STATE $e \leftarrow x_i - (x_r - X_i)$
\STATE $\Phi \leftarrow [x_i, u_i]$
\STATE $u \leftarrow$AdaptiveControl$(\Phi, e, \hat{\Theta}_i, P_i, B_i) + U_i$
\STATE $x_r \leftarrow x_r + F_r(x_r, u_r)\Delta t$
\STATE $x \leftarrow x + F(x, u)\Delta t$
\STATE $i \leftarrow$ $j : x, u \in S_j$
\ENDWHILE
\end{algorithmic}
\end{algorithm} 


\color{black}

\subsection{Nonaffine Systems}
We once again start with \eqref{eq:general_dynamics}, expand the dynamics around $(X_0,U_0)$, which together with Assumption 1 leads to a class of nonlinear systems 
\begin{equation}\label{eq:general_dynamics3}
\dot{x} = Ax + B[u + h(x,u)] 
\end{equation}
Noting that $u$ is required to minimize 
\eqref{eq:optimal_control}, we assume that $u$ can be expressed as an analytic function of the state $x$, which in turn leads to the assumption that
\be h(x,u)=g(x)\label{eq:nonaffine} \ee 
The goal is to determine $u$ in real time when parametric uncertainties are present in \eqref{eq:general_dynamics3}-\eqref{eq:nonaffine}  so that \eqref{eq:optimal_control} is accomplished for initial condition $x_0$. As before, the reference system dynamics with no parametric uncertainties 
\redc{is chosen as in \eqref{1b}. Noting that \eqref{eq:nonaffine} makes the plant  in \eqref{eq:general_dynamics3}  identical to \eqref{eq:affine}}, 
the same AC-RL controller in Section \ref{ss:acrl1} leads to global boundedness under Assumptions \ref{a1}-\ref{a3}, with valid extensions as in Sections \ref{sec:relax} and \ref{sec:MSAC-RL Controller}. We skip a formal statement of the result due to space limitations.

\section{Integrated AC-RL Solutions: Problem 2}\label{s:problem3}
We now consider Problem 2, which pertains to the class of dynamic system \eqref{equ:control_affine}, and introduce parametric uncertainties $\theta \in \mathbb{R}^{n_{\theta}}$ and $\nu \in \mathbb{R}^{n_{\nu}}$ in $f(X)$, and $\Lambda\in \mathcal{S}_m^+$ in $U$ which may be due to loss of effectiveness, leading to the system 
\begin{equation}\label{equ:extended_match_sys}
    \dot{X}(t) = f(X) - \rho(X) \nu + B(X)\left[\Lambda U(t) - \phi(X)\theta\right],
\end{equation}
where $\rho(X) \in \mathbb{R}^{n\times n_{\nu}}$ and $\phi(X) \in \mathbb{R}^{m \times n_{\theta}}$ are known nonlinear dynamics of the parametric uncertainties $\theta$ and $\nu$, respectively. It is clear that $\phi(X) \theta$ denotes a matched uncertainty similar to that introduced by Assumptions~\ref{a2}, while $\rho(X) \nu$ denotes an unmatched uncertainty. We make the following assumption regarding the latter, which is similar to an extended matching condition in \cite{Krstic1995}:
\begin{assumption}\label{assum:extended_match}
     The unmatched uncertainty satisfies $\rho(X) \nu \in span\{ad_f B\}$, where $ad_f B = [f, B] = \frac{\partial f}{\partial X} B - \frac{\partial B}{\partial X} f$ denotes the Lie bracket of $f, B$. $f(X)$ and $B(X)$ are nonlinear functions such that $B$ and $ad_f B$ are linearly independent.
\end{assumption}

For the system in \eqref{equ:extended_match_sys} with Assumption~\ref{assum:extended_match}, we now propose an AC-RL controller.  
We first consider a  control-affine reference system, similar to \eqref{equ:control_affine}, of the form,
\begin{equation}\label{equ:control_affine_ref}
    \dot{X}_r(t) = f(X_r) + B(X_r)U_r(t).
\end{equation}
Given an initial condition $X_0$, $U_r(t)$ minimizes the cost function given in \eqref{eq:optimal_control} where $x(t)$ and $u(t)$ are replaced by $X(t)$ and $U(t)$, respectively. The reference input $U_r(t)$ can be constructed using the offline RL-based approach in Section \ref{sec:prob-offline}. In addition to Assumption \ref{rl_policy} and \ref{a3}, we introduce the following assumption:

\begin{assumption}\label{assum:parameter_CCM}
    A uniformly bounded parameter-dependent contraction metric $M(X, \hat{\nu})$ can be computed for each parameter estimate $\hat{\nu}\in \mathbb{R}^{p_{\nu}}$.
\end{assumption}
Under Assumption~\ref{assum:parameter_CCM}, let $\gamma(t, s)$ be the geodesic that connects $X_r(t)$ to $X(t)$, i.e., $\gamma(t, 0) = X_r(t)$ and $\gamma(t, 1) = X(t)$. The Riemannian energy of the geodesic is defined as $E =  \int_0^1 \gamma_s^T M(X, \hat{\nu})\gamma_s ds$. We now describe the AC-RL controller in equations \eqref{equ:HTAC-prob3}-\eqref{equ:u_c} and the stability result:
\begin{subequations}\label{equ:HTAC-prob3}
    \begin{align}
        U &= \Psi \left(U_c + U_{\nu} + \phi(X) \hat{\theta} \right),\\
        U_{\nu} &= \sum_{i=1}^{p_{\nu}} \Omega_i \dot{\hat{\nu}}_i \int_0^1 \left[ \frac{\partial \rho_i^T(\gamma(s))}{\partial X_1}\ ...\ \frac{\partial \rho_i^T(\gamma(s))}{\partial X_{n-1}}\ 0 \right]\gamma_s(s) ds,\\
        \dot{\hat{\nu}} &= \beta_{\nu}(\nu_a - \hat{\nu}) \mathcal{N}_{\nu}, \\
        \dot{\nu}_a &= -\gamma_{\nu} \rho(X)^T M\gamma_s(1)E, \\
        \dot{\hat{\theta}} &= \beta_{\theta} (\theta_a - \hat{\theta})\mathcal{N}_{\theta}, \\
        \dot{\theta}_a &= -\gamma_{\theta} \phi(X)^T B(X)^T M \gamma_s(1)E,\\
        \dot{\Psi} &= \beta_{\Psi} (\Psi_a - \Psi) \mathcal{N}_{\Psi},\\
        \dot{\Psi}_a &= - \gamma_{\Psi} B(X)^T M \gamma_s(1) E U^T,
    \end{align}
\end{subequations}
where 
\begin{subequations}\label{equ:HTAC-reg}
    \begin{align}
        \mathcal{N}_{\nu} &= 1 + \omega_{\nu}^T\omega_{\nu}, \quad \omega_{\nu} = \rho(X)^T M \gamma_s(1),\\
    \mathcal{N}_{\theta} &= 1 + \omega_{\theta}^T \omega_{\theta}, \quad \omega_{\theta} = \phi(X)^T B(X)^T M \gamma_s(1),\\
    \mathcal{N}_{\Psi} &= 1 + \omega_{\Psi}^T \omega_{\Psi}, \quad \omega_{\Psi} = U,
    \end{align}
\end{subequations}
$\rho_i$ is the $i$th column of $\rho(X, t)$, $\Omega_i \in \mathbb{R}^m$ extracts the $j^{th}$ column vector of $B(x)$ such that $\rho_i(x) \nu_i \in span\{ ad_f b_j \}$, and $U_c$ is a nominal controller that leverages the RL-based input $U_r$ and satisfies 
\begin{align*}
    &\frac{\partial E}{\partial t} + 2\gamma_s(1)^T M(X, \hat{\nu}) (f(X) -\rho(X)^T\hat{\nu} + B(X) U_c )\\
    &- 2\gamma_s(0)^T M(X_r, \hat{\nu})\left[ f(X_r) + B(X_r) U_r \right] \leq -2\lambda E. \numberthis \label{equ:u_c}
\end{align*} where $\lambda > 0$. The adaptive gain parameters $\beta_{\nu}$, $\gamma_{\nu}$, $\beta_{\theta}$, $\gamma_{\theta}$, $\beta_{\Psi}$, $\gamma_{\Psi}$ are chosen as positive numbers that satisfy
\begin{equation}\label{equ:HTAC-P2-Gains}
    \frac{\beta_{\nu}}{\gamma_{\nu}} \geq \frac{3}{\lambda},\quad \frac{\beta_{\theta}}{\gamma_{\theta}} \geq \frac{3}{\lambda}, \quad \frac{\beta_{\Psi}}{\gamma_{\Psi}} \geq \frac{3\|\gamma_s(1)\|\|M\|\|B\|}{\lambda}
\end{equation}

\begin{theorem}\label{t:Contraction-HTAC}
    Under Assumption~\ref{rl_policy},~\ref{a3},~\ref{assum:extended_match}, and~\ref{assum:parameter_CCM}, the closed-loop system specified by the plant in \eqref{equ:extended_match_sys}, and the adaptive controller in \eqref{equ:HTAC-prob3}-\eqref{equ:HTAC-P2-Gains} has bounded solutions and ensures that 
    $\lim_{t\to\infty} \|X(t) - X_r(t)\| = 0$.
\end{theorem}

\begin{remark}
    Theorem~\ref{t:Contraction-HTAC} proposes a solution to Problem 2 and leverages contraction theory rather than Lyapunov theory. The new contributions here, in comparison with \cite{lopez2020adaptive}, are that the controller is based on AC-RL, it utilizes a high-order tuner, and incorporates larger uncertainties such as loss of effectiveness. 
\end{remark}

\section{Numerical Studies \remove{Validation}}\label{sec:num_val}
\subsection{Problem 1}
\label{sec:numerical}
\begin{figure}
	\centering
	\includegraphics[width=0.7\linewidth]{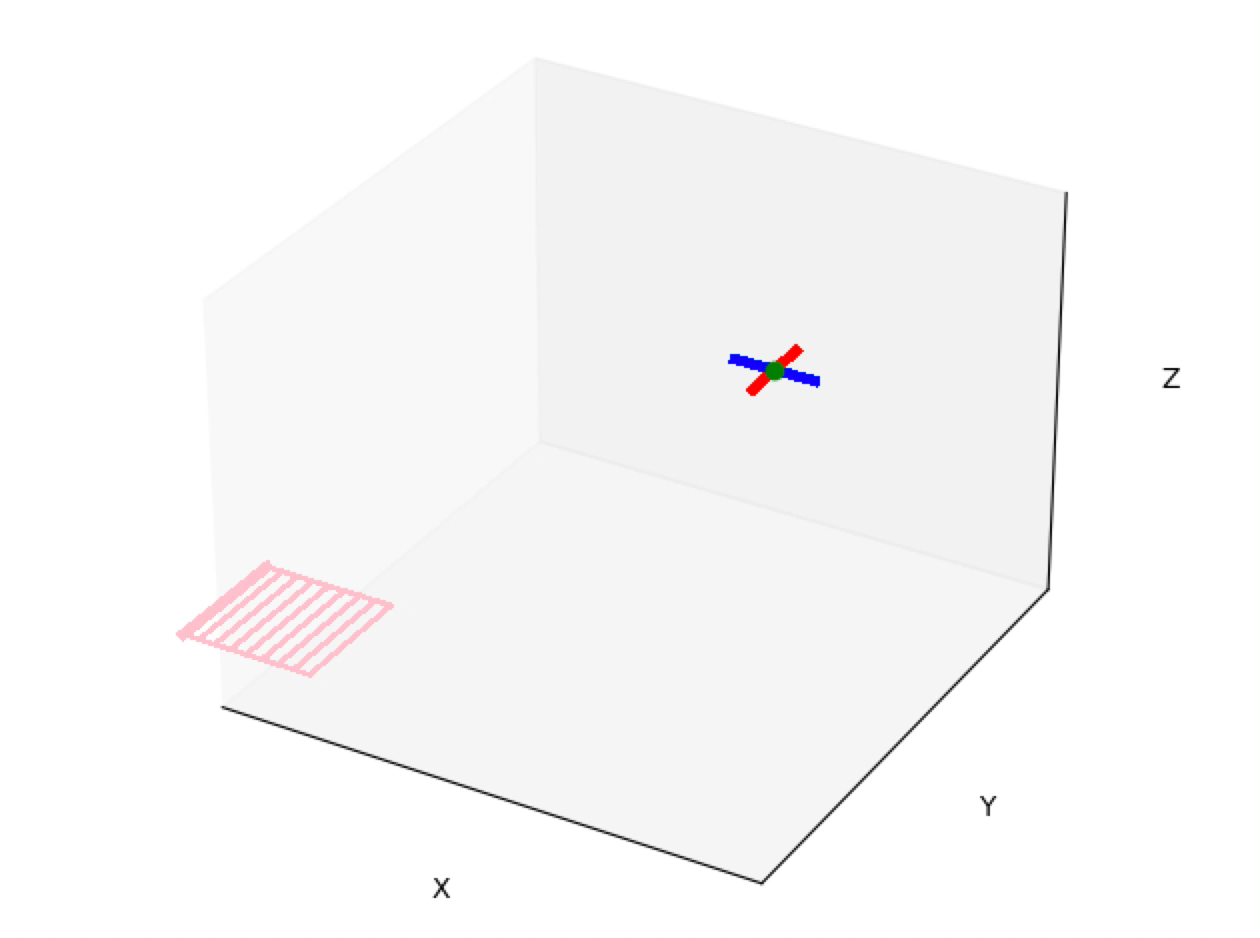}
	\caption{The task is to have a quadrotor (indicated with an X) to land on a moving platform (shown at the bottom left of the quadrant).}
	\label{fig:sim1}
\end{figure}

We consider a control task 
of a quadrotor moving in 3-D to land on a moving platform (See Figure \ref{fig:sim1}). 

The adaptive control design was based on a linearized quadrotor model
\begin{equation}\label{eq:linearized_dynamics}
	\begin{split}
		& \ddot{x} = g\theta \quad \ddot{\theta} = \frac{L}{I_y}\tau_\theta \\
		& \ddot{y} = -g\phi \quad \ddot{\phi} = \frac{L}{I_x}\tau_\phi\\
		& \ddot{z} = \frac{f_z - mg}{m} \quad \ddot{\psi} = \frac{1}{I_z}\tau_{\psi}
	\end{split}
\end{equation}
which defines the matrices $A_r, B$. {$\Theta_{l,r}$ was chosen via $\Theta_{l,r} = B^\dagger (A_r - A_H)$ such that all eigenvalues of $A_H$ were at $-6$, and we also chose $Q_H = I$.} The generalized forces were generated via the control input as:
\begin{equation}\label{eq:qr_moment_dynamics}
	\begin{aligned}
		\begin{bmatrix}
			f_z\\
			\tau_\phi\\
			\tau_\theta\\
			\tau_\psi\\
		\end{bmatrix} = 
		\begin{bmatrix}
			1 & 1 & 1 & 1\\
			L & 0 & -L & 0\\
			0 & L & 0 & -L\\
			\nu & -\nu & \nu & -\nu\\
		\end{bmatrix}\Lambda u
	\end{aligned}
\end{equation}
In \eqref{eq:linearized_dynamics},  $g = 9.81, m = 1.2, L = 0.3, I_x = I_y = 0.22, I_z = 0.44, \nu = 1.0$ (in SI units). $\Lambda$ was chosen to be diagonal with entries  decreased from the nominal case of unity to represent loss of effectiveness of each actuator. 
\remove{Because a linearized quadrotor model is used in the AC-RL control design, $g(\cdot)$ was set to $ = 0$. $Q_H = I^{12\times 12}$.} {A linearized quadrotor model was used in the AC-RL control design, and therefore $g(\cdot) = 0$ {in \eqref{1b}.}} A nonlinear quadrotor model as in \cite{Dydek12}, however, was used for the training of the RL as well as for the evaluation of the AC-RL controllers.

The cost function in \eqref{eq:optimal_control} was chosen as
\begin{equation}\label{eq:costfnc}
	\begin{aligned}
		c(\vec{x}, t) = 
		\begin{cases} 
			-1 & \texttt{box} \\
			1 & \neg \texttt{box} \wedge (\Delta z \leq 0 \vee t \geq T_{max})\\
			0 & \texttt{else}
		\end{cases}
	\end{aligned}
\end{equation}
where the boolean variable \texttt{box} is assumed to be True if ALL of the following simultaneously hold: $|\remove{\Delta} z| \leq 5{\rm cm}$, $|\Delta xy| \leq 25 {\rm cm}$, $|\phi| \leq 10^\circ$, $|\theta| \leq 10^\circ$, $|v_{xy}| \leq 50{\rm cm/s}$, $|{\dot{z}}\remove{v_z}| \leq 10 {\rm cm/s}$, {and} \brownc{$T_{{\rm max}}=10 {\rm sec}$}, where $|\Delta xy|^2=((x - X_{{\rm plat}})^2 + (y - Y_{{\rm plat}})^2)$, $v_{xy}=||[\dot x\;\dot y]^T||$, and $(X_{{\rm plat}},Y_{{\rm plat}})$ denote the platform position in the $(x,y)$-plane. $\vec{x}$ is the whole quadrotor state, and the integral in \eqref{eq:optimal_control} is computed with a discretization of $\Delta t=0.1$. $z=0$ was assumed to correspond to the plane of motion of the moving platform. Overall, \eqref{eq:costfnc} implies that the cost increases if the quadrotor leaves a compact set {$\mathcal{D}(\vec{x})$}, if its altitude falls below the platform altitude, or if the termination time exceeds $10 {\rm sec}$.

We use the popular PPO method \cite{schulman2017proximal} to determine the RL component  
even though PPO has been shown only to have properties of local optimality \cite{holzleitner2021convergence}, as it provides a powerful methodology for training. \brownc{Any departures of the state from the compact set are accommodated through suitable reward engineering, i.e. through the addition of  high magnitude penalties in the cost function only on out-of-set excursions.}
The following hyperparameters are used in the training of the RL algorithm, PPO (see \cite{schulman2017proximal} for details), in the outerloop: $\gamma = .99, \Delta t = .1, T = 1000, N = 64$. The fourth-order Runge-Kutta method is used to integrate the dynamics, with a step-size of $h = .001$. The learning rate used for SGD algorithms is $.0001$, and both the value and policy networks were chosen to have two hidden layers, each with $128$ neurons and ReLU activations. 
The inner-loop controller uses the MSAC algorithm described in \ref{sec:MSAC-RL Controller}, with $\gamma = 50, \beta = 25, \mu = 1$.



We test two types of parametric uncertainties: 1) the mass, length and inertia properties of the quadrotor varied between $\pm 25 \%$ of their nominal (reference) values (Table \ref{tab:table1}) and 2) an abrupt loss of effectiveness (LOE) in the fourth propeller  (Table \ref{tab:results1}) and therefore in $\Lambda$. 
Such a LOE may occur if the propeller blades are broken midflight, as demonstrated in \cite{Dydek12}. Both types of parametric uncertainties correspond to the target-system structure as in \eqref{1c}, with the symmetric part of $\Lambda$ being positive definite. {We observed that in all cases, the AC-RL ensured boundedness of all signals in the closed-loop system and that the tracking error converged to zero, which validates Theorem 4.}

\subsubsection{Results (Without Noise)}
{As our goal extends beyond Theorem 4\remove{, which is} to minimizing the cost in (57), we evaluate AC-RL using additional performance metrics, success rate (SR) and success time (ST), where}\remove{Two performance metrics, success rate (SR)  and success time (ST), are measured.} SR denotes the percentage of all trials where the quadrotor entered the compact set \remove{{$\mathcal{D}$}} {$\mathcal{D}(\vec{x})$} within 10 seconds and ST is the mean time required to complete this task over all successful tests. The results are reported in Tables \ref{tab:table1} and \ref{tab:results1}, with Table I also including a comparison of the average cost in \eqref{eq:costfnc} averaged over 1000 trials. {It should be noted that Theorem 4 provides no guarantees for SR to be at 100\% or for ST to be less than a specified amount, as both correspond to the transient performance of a nonlinear controller, which is very difficult to quantify. Nevertheless, the results reported in Tables \ref{tab:table1} and \ref{tab:results1} show that AC-RL outperforms its competitors.} \remove{The results are reported in Tables \ref{tab:table1} and \ref{tab:results1}, with Table I also including a comparison of the average cost in \eqref{eq:costfnc} averaged over 1000 trials.}

\begin{figure}
	\centering
	\includegraphics[width=\linewidth]{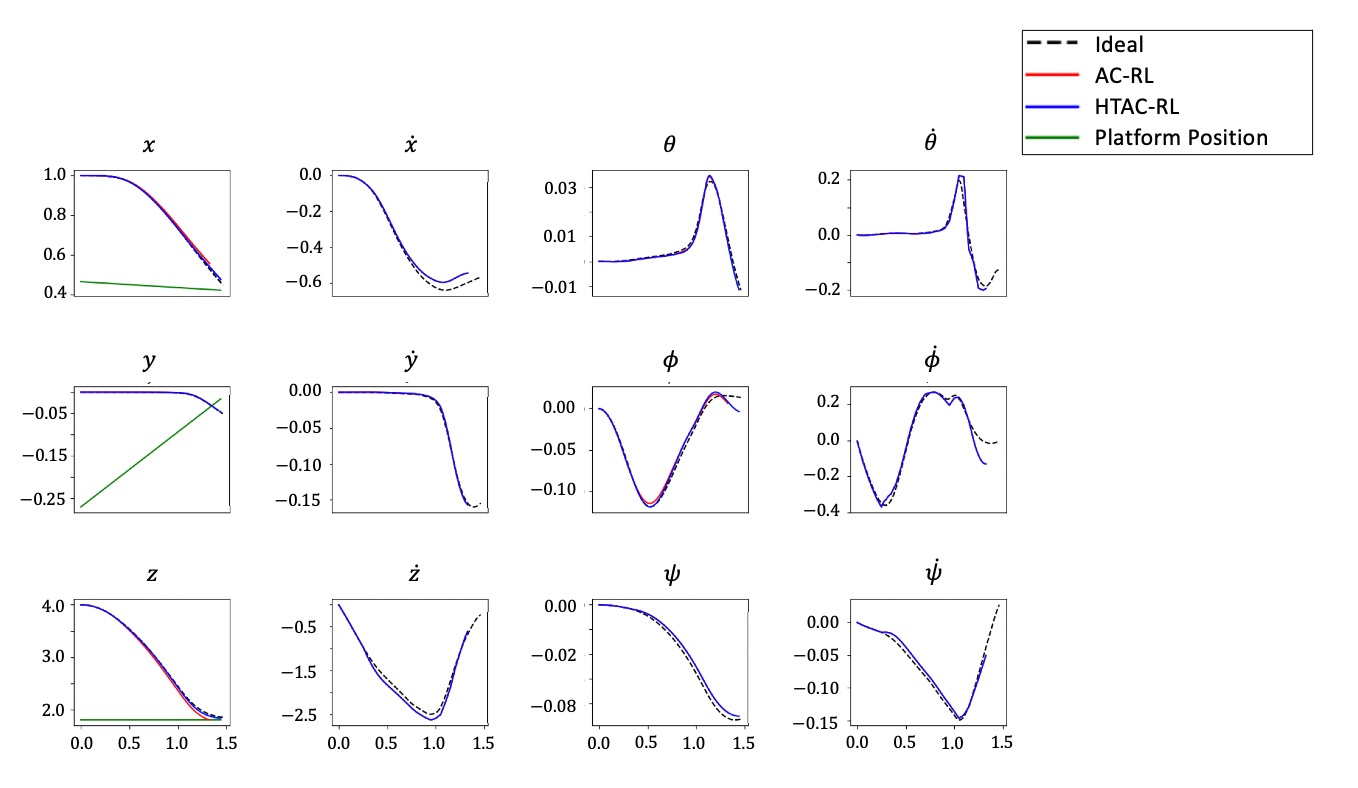}
	\caption{A comparison of the states of the quadrotor using the AC-RL and RL controllers when there is a symmetric 25\% parametric uncertainty. Also compared is the performance of the reference model with the RL controller with no parametric uncertainty (indicated as Ideal).}
	\label{fig:figsim2}
\end{figure}

\begin{figure}
	\centering
	\includegraphics[width=0.7\linewidth]{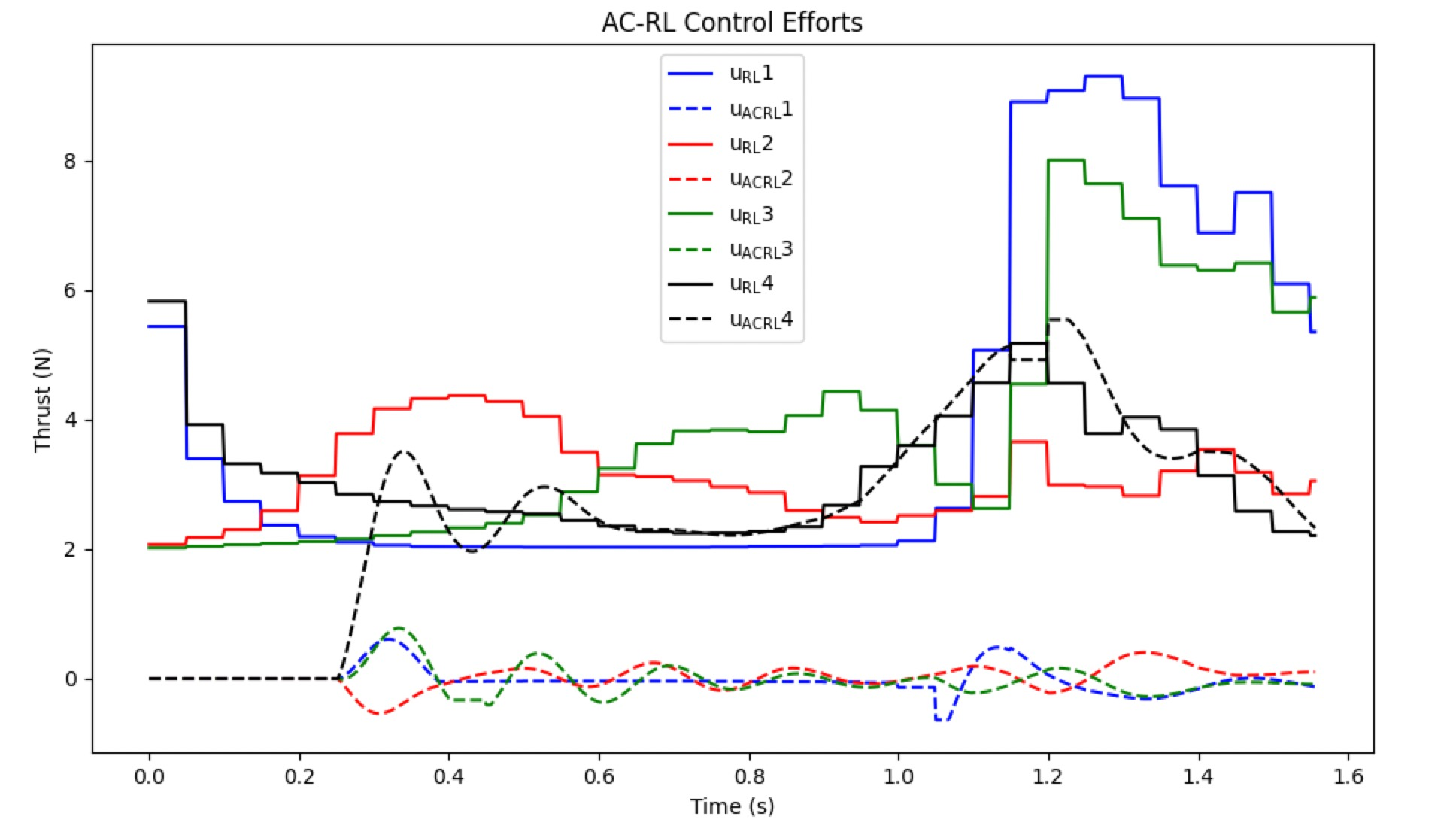}
	\caption{Control inputs corresponding to the AC-RL and RL controller, for the case of 50\% parametric uncertainty in motor 4.}
	\label{fig:figsim3}
\end{figure}

\begin{figure}
	\centering
	\includegraphics[width=0.7\textwidth]{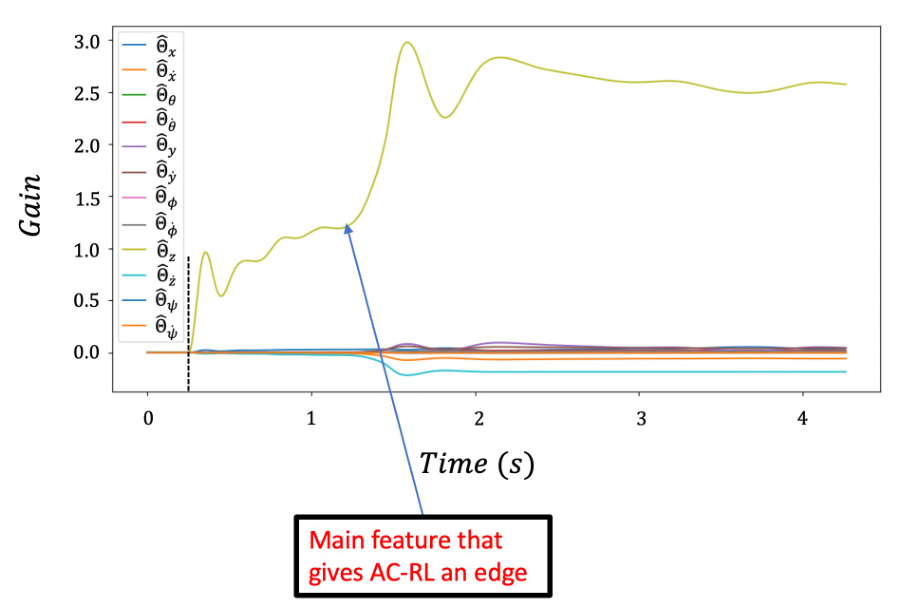}
	\caption{Adaptive control parameters for the same case shown in Figure \ref{fig:figsim3}.}
	\label{fig:figsim4}
\end{figure}

\begin{figure}
	\centering
	\includegraphics[width=\linewidth]{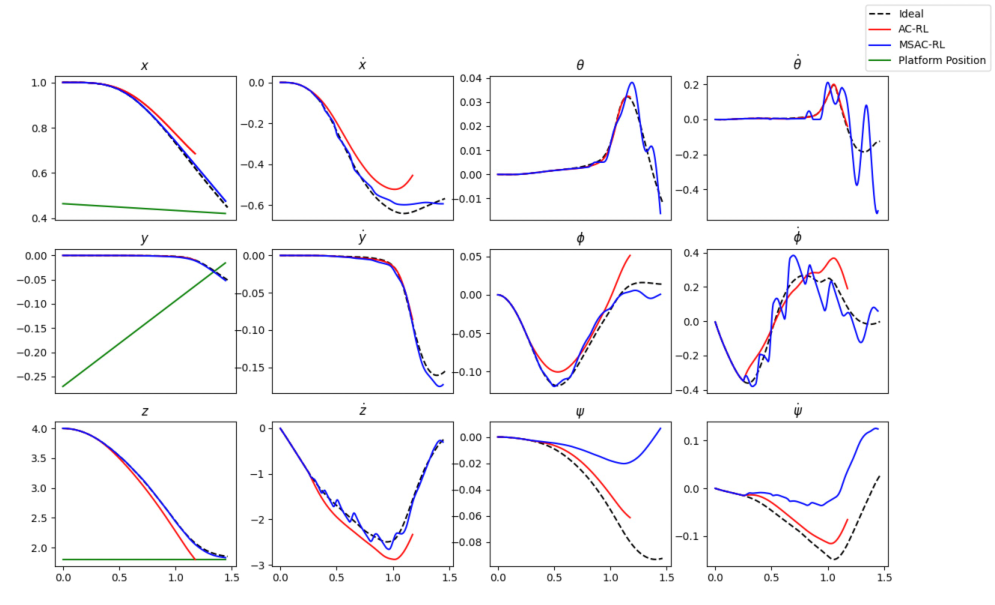}
	\caption{A magnitude-constraint of $\pm 10N$ is introduced in the control input for an asymmetric loss of effectiveness is incurred at 0.25 seconds, in motor 4. The resulting response with the MSAC-RL is shown.}
	\label{fig:figsim5}
\end{figure}

\begin{figure}
	\centering
	\includegraphics[width=0.7\linewidth]{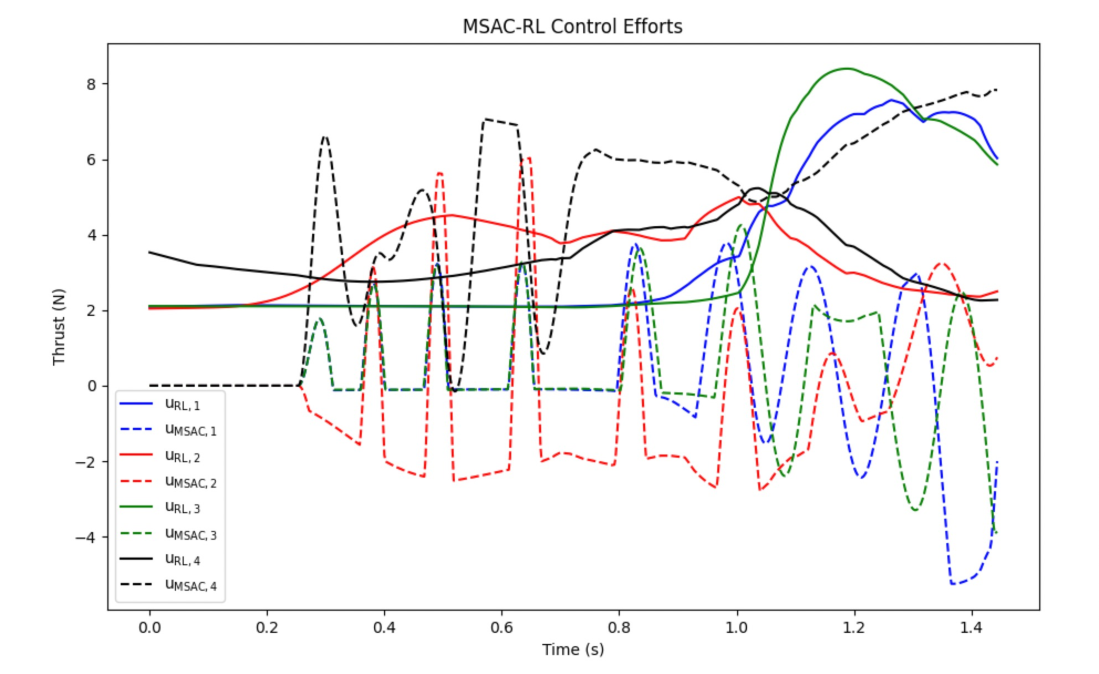}
	\caption{Plots of the RL policy inputs and the additional corrective inputs as calculated by the MSAC algorithm. An asymmetric loss of effectiveness is incurred at 0.25 seconds, which can be characterized as a form of parametric uncertainty. The magnitude limit in this example is 10 N (for all four inputs). The plot demonstrates that, when there is no uncertainty (e.g, no loss of effectiveness has been incurred), the adaptive controller does not alter the RL policy. Once the LOE occurs, the AC begins to generate additive inputs to accommodate the error between reference and true models.}
	\label{fig:figsim6}
\end{figure}

Figure \ref{fig:figsim2} shows all states corresponding to an ``ideal RL", with ``RL" and with ``AC-RL" controllers, where the latter two correspond to the case when there was a 20\% symmetric uncertainty of type 1). No magnitude constraint was introduced for the runs shown in this figure. The improved performance with AC-RL is obvious from the figure. The corresponding control inputs are shown in Figure \ref{fig:figsim3}. The change in the control input when there is an uncertainty is clearly visible from this figure, especially the motor inputs from $u_{ACRL3}$ and $u_{ACRL4}$.
In Figure \ref{fig:figsim4}, we show the corresponding parameters of the adaptive controller $\widehat\Theta$. In Figure \ref{fig:figsim5}, we show the responses of the state with MSAC-RL, where a magnitude constraint of $\pm 10N$ was introduced. The corresponding control efforts are shown in Figure \ref{fig:figsim6}.

We compare four control methods: 1) RL, which just uses the trained $\pi$ to dictate control; 2) AC-RL, which utilizes both $\pi$ and the adaptive laws in \ref{sec:MSAC-RL Controller}; 3) a meta-learning approach (ME-RL) which \bluec{employs} a deep-learning based policy update in addition to a network that learns all hyperparameters  \cite{finn2017metalearning};  and 4) a domain-randomized approach (DR-RL) \cite{Loquercio2019} which uses domain randomization of the plant parameters during policy training. The following observations follow:

\begin{table}[h!]
	\begin{center}
		\begin{tabular}{c |c  c c}
			\toprule 
			\multicolumn{1}{c|}{\textbf{Algorithm}} & \multicolumn{3}{c|}{\textbf{Results}}\\
			\toprule
			\textbf{} &  \textbf{SR} & 
			\textbf{ST} & \textbf{Cost \eqref{eq:costfnc}}\\
			\midrule 
			\textit{RL} &            $48\%$ &         $7.5s$ & 1.2\\
			\textit{AC-RL} & $        82\%$ & $3.5s$ & \orangec{-0.3}\\
			\textit{DR-RL} &            $75\%$ &         $7.1s$ & 0.11\\
			\textit{ME-RL} &            $88\%$ &         $3.4s$ & -0.41\\
			\bottomrule 
		\end{tabular}
		\caption{$\pm 25\%$ parametric uncertainty results}
		\label{tab:table1}
		
	\end{center}
	
\end{table}

\begin{enumerate}[(i)]
\item \remove{We note from Table \ref{tab:table1},} AC-RL performs better {than}\remove{when compared to} pure RL or DR-RL {(Table \ref{tab:table1})}. \\
\item \remove{We note that} ME-RL outperforms AC-RL on all three metrics. 
This comes with two qualifiers: 1) the meta-learner in ME-RL is trained using the same distributional shift on which it was tested (the $\pm 25 \%$ parameter perturbations), and 2) the meta-learner is a DNN {and thus has no guarantees of convergence, unlike AC-RL.} \remove{- hence the ME-RL policy does not provide the guarantees of convergence within a compact set that are afforded by AC-RL.} \\
\item The relevance of point 1) becomes apparent from the ME-RL results in Table \ref{tab:results1}, where it can be see that ME-RL greatly underperforms AC-RL. This is because the specific type of uncertainty (asymmetric LOE) studied in Table \ref{tab:results1} was not incorporated into the meta-learner's training regimen. The LOE column represents the degree of propeller thrust lost (with $0\%$ being no loss). For a 75\% LOE there is no data on the RL or ME-RL success time because there were no successful tests. \remove{No measurement noise was introduced in these experiments.} \\
\item {Table \ref{tab:table1} also shows that for a 25\% parametric uncertainty,} \remove{We have also carried out a comparison using the  cost metric in \eqref{eq:costfnc} for a 25\% parametric uncertainty (see Table I). We observed that} the average cost was  \orangec{-0.3} and 1.2 for AC-RL and RL, respectively. In comparison, the  average cost for the case when there were no uncertainties, was observed to be \orangec{-0.51} for a pure RL approach, which is the best achievable cost, i.e. AC-RL has an optimality gap of 40\%.  \\
\item {We observed that}\remove{We also carried out a cost comparison with a pure AC approach. Due to the stringent success metric in \eqref{eq:costfnc}, we found that a} pure AC led to success only with a smaller set of initial conditions for the state{, primarily due to the stringent metric in \eqref{eq:costfnc}}. This led to an optimality gap of \orangec{40\%} under AC-RL, in contrast to \orangec{115\%} under AC alone. \remove{This clearly illustrates that AC-RL is more optimal than AC alone.}
\end{enumerate}

\begin{table}[!ht]
	\begin{center}
		
		\begin{tabular}{c| c | c | c}
			\textbf{AC-RL} & \textbf{ME-RL} & \textbf{RL} & \textbf{LOE}\\
			\textbf{SR} & \textbf{SR} & \textbf{SR} & \textbf{}\\
			\midrule %
			$97\%$ & $98\%$ & $97\%$ & $0\%$\\
			$88\%$ & $78\%$ & $74\%$ & $10\%$\\
			$73\%$ & $41\%$ & $34\%$ & $25\%$\\
			$54\%$ & $8\%$ & $14\%$ & $50\%$\\
			$27\%$ & $0\%$ & $0\%$ & $75\%$\\
			\midrule %
			\textbf{AC-RL} & \textbf{ME-RL} & \textbf{RL} & \textbf{LOE}\\
			\textbf{ST} & \textbf{ST} & \textbf{ST} & \textbf{}\\
			\midrule %
			$2.6s$ & $2.8s$ & $2.6s$ & $0\%$\\
			$3.1s$ & $3.9s$ & $3.1s$ & $10\%$\\
			$2.9s$ & $5.1s$ & $4.7s$ & $25\%$\\
			$3.3s$ & $7.5s$ & $5.4s$ & $50\%$\\
			$3.3s$ & $--$ & $--$ & $75\%$\\
		\end{tabular}
		\caption{Results from the simulated quadrotor experiments. }
		\label{tab:results1}
	\end{center}
\end{table}

\subsubsection{Results (with noise)}
We further studied (experimentally) the efficacy of AC-RL when noise is introduced into the system. A set of new reference environments are constructed, in which aleatoric measurement noise is introduced. Two reference environments are used - one with positional measurement noise only, and one with positional and orientation measurement noise. In each case the noise is taken to be zero mean and normally distributed - this may correspond, for example, to posterior position/orientation estimates that result from a Kalman filtering algorithm. RL is used to train a policy for each of these environments, and the AC-RL and RL approaches are tested on the quadrotor task with LOE and the appropriate measurement noise. The results are reported in Tables \ref{tab:results-noise} and \ref{tab:results-noise2}, which clearly indicate the large improvement of AC-RL over RL.

\begin{table}[!ht]
	\centering
	\begin{tabular}{c| c | c | c}
		\textbf{AC-RL} & \textbf{ME-RL} & \textbf{RL} & \textbf{LOE}\\
		\textbf{SR} & \textbf{SR} & \textbf{SR} &  \\
		\midrule %
		
		$93\%$ & $89\%$ & $93\%$ & $0\%$\\
		$68\%$ & $77\%$ & $42\%$ & $25\%$\\
		$31\%$ & $35\%$ & $4\%$ & $50\%$\\
	\end{tabular}
	\caption{Success rates for LOE with position measurement noise. Measurement of the Cartesian positions $x, y, z$ contain additive aleatoric uncertainty given by random variables drawn from $\mathcal{N}(0, 0.05)$.}
	\label{tab:results-noise}
\end{table}

\begin{table}[!ht]
	\centering
	
	\begin{tabular}{c| c | c | c}
		\textbf{AC-RL} & \textbf{ME-RL} & \textbf{RL} & \textbf{LOE}\\
		\textbf{SR} & \textbf{SR} & \textbf{SR} &  \\
		\midrule %
		
		$84\%$ & $79\%$ & $84\%$ & $0\%$\\
		$51\%$ & $33\%$ & $24\%$ & $25\%$\\
		$23\%$ & $12\%$ & $0\%$ & $50\%$\\
	\end{tabular}
	
	\caption{Success rates for LOE with position and orientation measurement noise. Measurement of the Cartesian positions $x, y, z$ contain additive aleatoric uncertainty given by random variables drawn from $\mathcal{N}(0, 0.05)$ and $\phi, \theta, \psi$ are perturbed by random variables drawn from $\mathcal{N}(0, 5 \frac{\pi}{365})$.}
	\label{tab:results-noise2}
\end{table}

\subsection{Problem 2}
\begin{figure}
	\centering
	\includegraphics[width=0.7\textwidth]{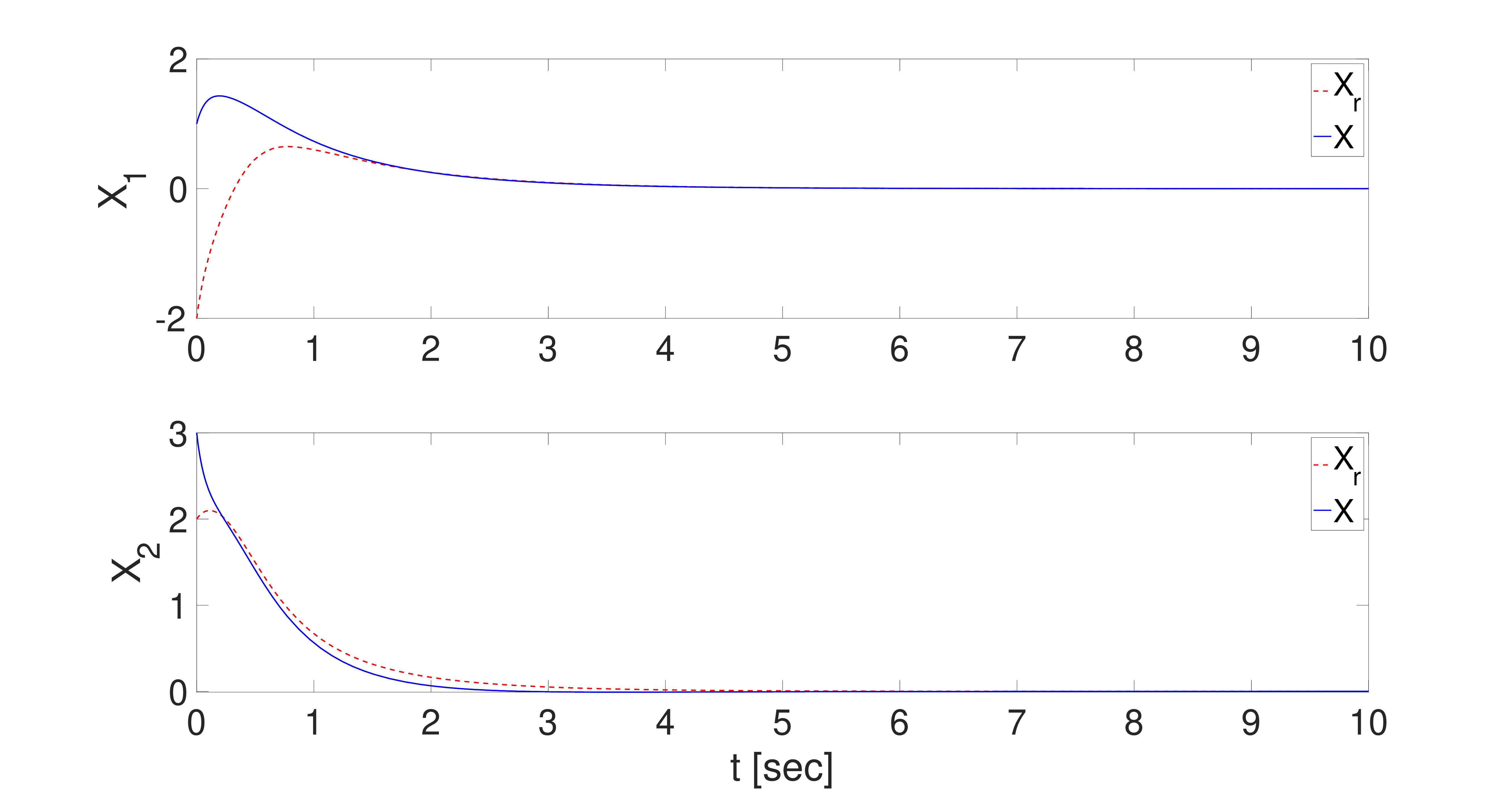}
	\caption{State trajectories of reference and true systems.}
	\label{fig:problem2}
\end{figure}

\bluec{To demonstrate the effectiveness of our proposed AC-RL controller in \eqref{equ:HTAC-prob3}-\eqref{equ:HTAC-P2-Gains}, we provide numerical simulation results for a system described in \eqref{equ:extended_match_sys} with
	\begin{align*}
		X &= \begin{bmatrix}
			X_1 \\ X_2
		\end{bmatrix}, f(X) = \begin{bmatrix}
			- X_1 - X_1^3 + X_2^2 \\ 0
		\end{bmatrix}, B(X) = \begin{bmatrix}
			0 \\ 1
		\end{bmatrix},\\
		\phi(X) &= X^T, \rho(X) = \begin{bmatrix}
			0 & 0
		\end{bmatrix}^T, \Lambda = I , \theta = \begin{bmatrix}
			1 & 1
		\end{bmatrix}^T.
	\end{align*}
	Note that unmatched uncertainty $\rho(X)\nu$ in \eqref{equ:extended_match_sys} is removed for simplicity in the numerical example. As a result, the contraction metric $M(X)$ does not depend on parameter estimate $\hat{\nu}$. The initial estimates $(\hat{\theta}, \theta_a)$ and $(\Psi, \Psi_a)$ for unknown parameters $\theta$ and $\Lambda$, respectively, are 20\% off of true values. As shown in Figure~\ref{fig:problem2}, our proposed controller provides bounded closed-loop solution $X(t)$ that converges to reference state trajectory $X_r(t)$ generated using RL-based controller $U_r$ asymptotically, which validates theoretical results stated in Theorem~\ref{t:Contraction-HTAC}. }

\section{Summary and Conclusions}
This paper proposed solutions for real time control and learning in dynamic systems {with parametric uncertainties} using a combination of adaptive control and reinforcement learning. Two classes of nonlinear dynamic systems are considered, both of which are control-affine. The first class of dynamic systems utilizes equilibrium points and expansion forms around these points and employs a Lyapunov approach. The second class of nonlinear systems uses
contraction theory as the underlying framework. For both classes of systems, the AC-RL controller is shown to lead to online policies that guarantee stability, and leverage accelerated convergence properties using a high-order tuner. {In both cases, when there are no uncertainties, the RL component is assumed to be capable of generating policies that are guaranteed to be stable and near-optimal.} Additionally, for the first class of systems, the AC-RL controller is shown to lead to parameter learning with persistent excitation. Together, this paper takes a first step towards real-time control using machine learning with provable guarantees, by drawing upon key insights, tools, and approaches developed in these two disparate and powerful methodologies of adaptive control and reinforcement learning. Several more steps need to be carried out in this direction and address the gap between stability and optimality of general nonlinear systems with arbitrary cost functions.

\label{sec:sum}

\section{Appendix}
\subsection{Proof of Theorem \ref{th:classic}}
\label{Proofth1}
	A Lyapunov function candidate is chosen as
	\begin{equation}\label{eq:Lyap1}
		\begin{split}
			V(e, \widetilde{\Theta})
			&= e^TPe+\tr\left(\widetilde{\Theta}^T\left(\Lambda ^{T} S\right)\widetilde{\Theta}\right)
	\end{split}\end{equation}
	where  $S=\Gamma^{-1}$ in \eqref{eq:gradadaptive} with the symmetric part of $\Lambda S$  positive-definite. Using arguments as in \cite{Som10}, we can show that $\dot V =-e^TQ_a e$ 
	by using \eqref{eq:errormodel} and \eqref{eq:gradadaptive}. Thus, $e(t)$ and the parameter estimates $\widehat\Theta$ are uniformly bounded. If in addition $\Phi(t)$ is bounded, we note that $\dot{e}(t)$ is bounded, and that $e\in{\cal L}_2$. 
	It follows from Barbalat's lemma that $\lim_{t \to \infty}\|e(t)\| = 0$. 

\subsection{Proof of Theorem \ref{t:HAC-RL}}
	The closed-loop system dynamics with the AC-RL controller is represented by the error equation \eqref{52}, the adaptive control input \eqref{eq:adaptive_control}, and the adaptive laws in \eqref{eq:adaptive_param1}-\eqref{eq:mu}.
	A Lyapunov function candidate
	\begin{equation}
		\label{eq:lyap}
			V = \frac{1}{\gamma}\tr\left[(\Xi - \Theta)^T {\Lambda}^T (\Xi - \Theta)\right] + \frac{1}{\gamma}\tr\left[(\widehat\Theta - \Xi)^T {\Lambda}^T (\widehat\Theta - \Xi)\right] + e^T Pe
	\end{equation}
	yields a time-derivative
		\begin{eqnarray}
			\dot{V} 
			&=& -\tr\left[(\Xi - \Theta)^T(\Lambda + {\Lambda}^T) B^T Pe\Phi^T\right] - \frac{\beta}{\gamma} \tr\left[(\widehat\Theta - \Xi)^T(\Lambda + {\Lambda}^T)(\widehat\Theta - \Xi)\right] \N_t\nonumber\\
			&& + \tr\left[(\widehat\Theta - \Xi)^T (\Lambda + {\Lambda}^T) B^T Pe\Phi^T\right] + e^T\left(A_H^T P + PA_H\right)e + 2e^T PB\Lambda\widetilde{\Theta}\Phi\label{idealVdot}
		\end{eqnarray}

	Through algebraic manipulations, it can be shown that
	\begin{align}
		\dot V
		&\leq -\frac{\beta}{\gamma}\tr\left[(\widehat\Theta - \Xi)^T\Omega^T\Omega(\widehat\Theta - \Xi)\right] - \frac{\beta}{\gamma}\tr\left[(\widehat\Theta - \Xi)^T\Omega^T\Omega(\widehat\Theta - \Xi)\right]\mu\|\Phi\|^2 \\
		&\quad - 2\|e\|^2 + 4e^TPB\Lambda(\widehat{\Theta} - \Xi)\Phi \\
		&\leq - \frac{2\beta}{\gamma} \tr\left[(\widehat\Theta - \Xi)^T\Omega^T\Omega(\widehat\Theta - \Xi)\right] - 2\|e\|^2 \nonumber \\
		&\quad -4\left\|PB\right\|_2^2 \tr\left[(\widehat\Theta - \Xi)^T\Omega^T\Omega(\widehat\Theta - \Xi)\right]\|\Phi\|^2 \nonumber \\
		&\quad + 4\|e\| \|PB\|_2 \|\Lambda(\widehat\Theta - \Xi)\|_F \|\Phi\| \nonumber \\
		&\leq - \|e\|^2 - \frac{2\beta}{\gamma} \left\|\Omega(\widehat\Theta - \Xi)\right\|_F^2 -\left[\|e\| - 2\|PB\|_2 \brownc{\|\Omega\|_2} \|\Omega(\widehat\Theta - \Xi)\|_F \|\Phi\|\right]^2\leq 0 
		\label{eqn:vdot_leq_0}
	\end{align}
	where we have expressed the positive definite matrix $\Lambda$ using $\Omega$ with \brownc{$2\Omega^T\Omega = \Lambda+\Lambda^T$}, $\|\cdot\|_F$ is the Frobenius matrix norm and $\|\cdot\|_2$ is the matrix 2-norm. In the above derivation we have used \eqref{eq:4}, \eqref{eq:mu}, Assumption \ref{a3}, and the inequality $\|AB\|_F\leq\|A\|_2\|B\|_F$. It can be concluded that 
	$e,\widehat\Theta,\Xi\in\mathcal{L}_\infty$. From Assumption 2 and \eqref{53}, we conclude that $\Phi$ is bounded. Using Barbalat's lemma and arguments as in Appendix \ref{Proofth1}, we conclude that $\lim_{t\rightarrow\infty} e(t)=0$.
	
	\subsection{Proof of Corollary \ref{cor:cost}}
	\bluec{From Theorem \ref{t:HAC-RL}, it follows that $\lm{t}e(t) \to 0$. Using \eqref{52}, it follows that 
		\begin{equation} \label{eqn:conv_integral}
			\lm{t} \int_{t_0}^t \exp(A_H(t - \tau))B\Lambda[u(\tau) - \Theta\Phi(\tau)]d\tau = 0.
		\end{equation}
		It also follows that $\Phi \in \mathcal{L}_\infty$. Additionally, from the proof of Theorem \ref{t:HAC-RL} we know that $\widehat{\Theta}, \Xi \in \mathcal{L}_\infty$. Using \eqref{eq:adaptive_control} and \eqref{eq:adaptive_param3} and the supposition that $\dot{\Phi} \in \mathcal{L}_\infty$, we have $u, \dot{u} \in \mathcal{L}_\infty$. From here, with \eqref{eqn:conv_integral} it can be shown using first principles of real analysis that $\lm{t} [u(t) - \Theta\Phi(t)] = 0$, implying conclusion (i).} \redc{Due to the continuity arguments on $c(x, u)$ as noted in Section \ref{sec:prob1}, 
		we have that $\lm{t} [c(x, u) - c(x_r, u^*)] = 0$.}

	\subsection{Proof of Lemma \ref{lemma:1}}
	Note: In the proofs of Lemma \ref{lemma:1} - \ref{lemma:3} and Theorem \ref{theo:1}, we assume that $\|B_0\Phi(t)^T\| \leq c_1$. Such a $c_1$ exists as $\Phi$ is bounded following Theorem \ref{t:HAC-RL}.
	
		Consider the following candidate Lyapunov function
		\begin{equation*}
			V = \frac{\Lambda}{\gamma}\|\tilde{\vartheta}\|^2 + \frac{\Lambda}{\gamma}\|\tilde\theta - \tilde\vartheta\|^2 + e^T Pe
		\end{equation*}
		Similar to what has been shown in the proof of Theorem \ref{t:HAC-RL}, the time derivative of $V$ may be bounded by
			\begin{align*}
				\dot{V} &\leq -\frac{2\beta\Lambda}{\gamma}\|\tilde\theta - \tilde\vartheta\|^2 - \Lambda\|e\|^2 
				\leq -c_2\|x_1\|^2
			\end{align*}
			where $c_2 = \Lambda \min\left\{1, \frac{2\beta}{\gamma}\right\}$.
		Noting that $P$ is a positive definite matrix, it follows that for any vector $v\in\R^n$, $\alpha,\rho>0$ exist such that 
		\begin{equation}
			\label{eq:100}
			\alpha v^T v \leq v^T P v \leq \rho v^T v
		\end{equation}
		We prove Lemma \ref{lemma:1} by using contradiction. Assume $\mu(S) > n$ and denote $\overline{S} = \left\{t\in[t_1, \infty)\right\}$. Integrating $\dot V$, we have
		\begin{align*}
			\int_{t_1}^\infty \dot V(\tau) d\tau &= \int_S\dot V(\tau) d\tau + \int_{\bar{S} - S} \dot V(\tau) d\tau\\
			&\leq \int_S- c_2 \|x_1(\tau)\|^2 d\tau + \int_{\overline{S} - S}\dot V(\tau) d\tau \\
			&\leq -c_2 n\epsilon_2^2
		\end{align*}
		Choose $n(\epsilon_1, \epsilon_2) = c_3\epsilon_1^2 / (c_2\epsilon_2^2)$. This leads to a contradiction since $\|V(t_1)\| \leq c_3 \epsilon_1^2$, where $c_3 = \max\left\{\frac{\Lambda}{\gamma}, \rho \right\}$.
	
	\subsection{Proof of Lemma \ref{lemma:2}}
		Let $z(t)$ be a solution with initial condition $\|z(t_1)\|\leq\epsilon_1$. Suppose that $\left\|\tilde{\vartheta}(t)\right\|\geq \delta$ for all $t\in[t_1, t_1 + T]$, where $T = T_0 + \delta_0$.
		
		From the error model in \eqref{eq:34}, for any $t \geq t_1$,
		\begin{equation}
			\label{eq:47}
			e(t + \delta_0) = e(t) + \int_{t}^{t+\delta_0}A_H e(\tau) + B_0\Phi(\tau)^T\tilde{\theta}(\tau)d\tau
		\end{equation}
		from which we have
		\begin{align}
			\|e(t + \delta_0)\| \geq \left\|\int_t^{t + \delta_0}B_0\Phi(\tau)^T\tilde{\theta}(\tau)d\tau\right\| - \left\| e(t) + \int_t^{t+\delta_0}A_H e(\tau)d\tau\right\|
			\label{eq:48}
		\end{align}
		Given $\epsilon' = \epsilon_0\delta / (2c_2\delta_0)$ and $T = T_0 + \delta_0$, from the adaptive law in \eqref{eq:adp2}, there is an $\epsilon_2 > 0$ such that if $z(\tau)$ is a solution to \eqref{eq:adp1}-\eqref{eq:adp2} with $\|x_1(\tau)\| \leq \epsilon_2$ for all $\tau \in [t_1, t_1 + T]$, then $\|\tilde{\theta}(\tau) - \tilde{\theta}(t_1)\| \leq \epsilon'$.
		Define $\epsilon = \min\left\{\frac{\delta\epsilon_0}{8}, \frac{\delta\epsilon_0}{8c_1\delta_0}, \epsilon_2\right\}$. Now we show the lemma holds for this choice of $T$ and $\epsilon$.
		
		If $\|x_1(t_2)\| \geq \epsilon$ for some $t_2 \in[t_1, t_1 + T]$, then we are done. Assume $\|x_1(t)\| \leq \epsilon$ for all $t\in [t_1, t_1 + T]$, then 
		\begin{equation*}
			\left\|e(t) + \int_t^{t+ \delta_0}A_H e(\tau)d\tau\right\| \leq \epsilon + c_1\epsilon\delta_0 \leq \frac{\epsilon_0 \delta}{4}
		\end{equation*}
		for all $t\in[t_1, t_1 + T]$.
		By hypothesis, there exist a $t'\in[t_1, t_1 + T_0]$ such that
	\begin{equation*}
		\left\|\int_{t'}^{t' + \delta_0}B_0\Phi(\tau)^T wd\tau\right\| \geq \epsilon_0
	\end{equation*}
	where $w = \tilde{\theta}(t_1) / \left\|\tilde{\theta}(t_1)\right\|$ is a unit vector.
	
	We have
	\begin{align*}
		\left\|\int_{t'}^{t' + \delta_0}B_0\Phi(\tau)^T\left[w\left\|\tilde{\theta}(t_1)\right\| - \tilde{\theta}(\tau)\right]d\tau\right\| \leq c_2\int_{t'}^{t' + \delta_0}\left\|\tilde{\theta}(t_1) - \tilde{\theta}(\tau)\right\|d\tau\leq c_2\delta_0 \epsilon'= \frac{\epsilon_0\delta}{2}
	\end{align*}
	Since $\|e(\tau)\| \leq \|x_1(\tau)\| \leq \epsilon \leq \epsilon_2$ for $\tau \in[t_1, t_1 + T]$,
	\begin{align*}
		\left\|\tilde{\theta}(t_1)\right\|\left\|\int_{t'}^{t' + \delta_0}B_0\Phi(\tau)^T wd\tau\right\| - \left\|\int_{t'}^{t' + \delta_0}B_0\Phi(\tau)^T \tilde{\theta}(\tau)d\tau\right\| \leq \frac{\epsilon_0\delta}{2}
	\end{align*}
	which implies
	\begin{equation*}
		\left\|\int_{t'}^{t' + \delta_0}B_0\Phi(\tau)^T \tilde{\theta}(\tau)d\tau\right\| \geq \epsilon_0\delta - \frac{\epsilon_0\delta}{2} = \frac{\epsilon_0\delta}{2}
	\end{equation*}
	Thus
	\begin{equation*}
		\|x_1(t' + \delta_0)\| \geq \|e(t' + \delta_0)\| \geq \frac{\epsilon_0\delta}{2} - \frac{\epsilon_0\delta}{4} = \frac{\epsilon_0\delta}{4} > \epsilon
	\end{equation*}
	which is a contradiction.

\subsection{Proof of Lemma \ref{lemma:3}}
	By Lemma \ref{lemma:2}, the assumption that $\|z(t_1)\|\leq \epsilon_1$ and $\left\|\tilde{\vartheta}(t_2)\right\| \geq \delta$ implies that there exist an $\epsilon$ such that $\|x_1(t)\|$ is periodically both less than $\epsilon / 2$ and greater than $\epsilon$. This leads to a contradiction with Lemma \ref{lemma:1} if we choose $\epsilon_1 = \|z(t_1)\|$ and $\epsilon_2 = \epsilon / 2$. We thus conclude that $\left\|\tilde{\vartheta}(t_2)\right\| \leq \delta$. 

\subsection{Proof of Theorem \ref{theo:1}}
	Consider the candidate
	\begin{equation}
		\label{eq:40}
		V = \frac{\Lambda}{\gamma}\|\tilde\vartheta\|^2 + \frac{\Lambda}{\gamma}\|\tilde\theta - \tilde\vartheta\|^2 + e^T Pe
	\end{equation}
	With $\mu \geq 2\gamma\|PB\|^2/\beta$ and $Q_H \geq 2 I$ which solves $A_H^T P + PA_H = -Q_H$, the time derivative of \eqref{eq:40} may be bounded by
	\begin{equation}
		\label{eq:38}
		\dot{V} \leq -\frac{2\beta\Lambda}{\gamma}\|\tilde\theta - \tilde\vartheta\|^2 - \Lambda\|e\|^2 \leq 0
	\end{equation}
	Since $P$ is positive-definite, \eqref{eq:100} holds for some $\alpha,\rho>0$.
	Now we show that given $\epsilon_1 > \epsilon_2 > 0$, there is a $\eta$ with $0 < \eta < 1$ and $\Delta T_1 > 0$ such that if $z(t)$ is a solution with
	\begin{equation*}
		\epsilon_2 \leq V(t) \leq \epsilon_1, \quad\mathrm{for}\; t\in[t_1, t_1 + \Delta T_1],
	\end{equation*}
	then there is a $t_2\in[t_1, t_1 + \Delta T_1]$ such that $V(t_2) \leq \eta V(t_1)$. 
	Choose $0 < \nu < 1$, $\nu \leq \sigma < 1$ and $\Delta T_2 > 0$ so that $\rho\sqrt{1 - \sigma} - \Delta T_2\left(\frac{c_1}{\sqrt{\alpha}} + 2c_2\sqrt{\gamma}\right) > 0$, $\allowbreak \sqrt{\gamma(1 - \nu)} - \Delta T_2\left(\beta\sqrt{\gamma} + \frac{c_2\gamma\rho}{\sqrt{\alpha}}\right) > 0$, $0 < \Delta T_2 \big[\rho\sqrt{1 - \sigma} - \Delta T_2 \left(\frac{c_1}{\sqrt{\alpha}} + 2c_2 \sqrt{\gamma}\right)\big]^2 < 1$ and $0 <  \frac{2\beta\Delta T_2}{\gamma}\big[\sqrt{\gamma(1 - \nu)} - \Delta T_2\left(\beta\sqrt{\gamma} + \frac{c_2\gamma\rho}{\sqrt{\alpha}}\right)\big]^2 < 1$. From Lemma \ref{lemma:3}, we can obtain a $T$ when $\epsilon = \epsilon_1$ and $\delta = \sqrt{\epsilon_2 \nu}$. Define $\eta = 1 - \min\bigg\{\Delta T_2 \Big[\rho\sqrt{1 - \sigma} - \Delta T_2\left(\frac{c_1}{\sqrt{\alpha}} + 2c_2\sqrt{\gamma}\right)\Big]^2, \allowbreak \frac{2\beta\Delta T_2}{\gamma}\left[\sqrt{\gamma(1 - \nu)} - \Delta T_2\left(\beta\sqrt{\gamma} + \frac{c_2\gamma\rho}{\sqrt{\alpha}}\right)\right]^2\bigg\}$ and $\Delta T_1 = T + \Delta T_2$. Next we show that for this $\eta$ and $\Delta T_1$ the results hold.
	
	Let $t_2' \in [t_1, t_1 + T]$ be such that $\left\|\tilde{\vartheta}(t_2')\right\| \leq \delta\sqrt{\gamma}$. If $V(t_2')\leq \epsilon_2$, we are done. If $V(t_2') \geq \epsilon_2$, then
	\begin{equation}
		\label{eq:43}
		V(t_2') = \frac{\Lambda}{\gamma}\left\|\tilde{\vartheta}(t_2')\right\|^2 + \frac{\Lambda}{\gamma}\left\|\tilde{\theta}(t_2') - \tilde{\vartheta}(t_2')\right\|^2 + e(t_2')^T Pe(t_2')
	\end{equation}
	implies
	\begin{align}
		\label{eq:44}
		(1 - \nu)V(t_2') &\leq V(t_2') - \delta^2\\
		&\leq \frac{\Lambda}{\gamma}\left\|\tilde{\theta}(t_2') - \tilde{\vartheta}(t_2')\right\|^2 + e(t_2')^T Pe(t_2')\nonumber \\
		&\leq \frac{\Lambda}{\gamma}\left\|\tilde{\theta}(t_2') - \tilde{\vartheta}(t_2')\right\|^2 + \rho e(t_2')^2 \nonumber
	\end{align}
	Case 1: $\frac{\Lambda}{\gamma}\|\tilde{\theta}(t_2') - \tilde{\vartheta}(t_2')\|^2 < (1 - \nu)V(t_2')$. From \eqref{eq:44},
	\begin{equation}
		\label{eq:46}
		e(t_2')^2 \geq \frac{1}{\rho}(1 - \sigma)V(t_2'),
	\end{equation}
	where $0 < \nu \leq \sigma < 1$.
	From \eqref{eq:34}, for any $t \geq t_2'$,
	\begin{align*}
		\|e(t_2')\| - \|e(t)\| &\leq 
		\int_{t_2'}^t \left\|A_H e(\tau) + B_0\Phi(\tau)^T\tilde{\theta}(\tau)\right\|d\tau\\
		&\leq \left(\frac{c_1}{\sqrt{\alpha}} + 2c_2\sqrt{\gamma}\right)(t - t_2')\sqrt{V(t_2')}
	\end{align*}
	where the last inequality is due to the assumption that $\|A_H\|\leq c_1$ and $\left\|B_0\Phi(\tau)^T\right\|\leq c_2$ for all $\tau$.
	If $t_2 = t_2' + \Delta T_2$, we obtain
	\begin{align*}
		\|e(t)\| &\geq \|e(t_2')\| - \left(\frac{c_1}{\sqrt{\alpha}} + 2c_2\sqrt{\gamma}\right)(t_2 - t_2')\|z(t_2')\|\\
		&\geq \rho\sqrt{1 - \sigma}\sqrt{V(t_2')} - \left(\frac{c_1}{\sqrt{\alpha}} + 2c_2\sqrt{\gamma}\right)\Delta T_2\sqrt{V(t_2')}\\
		&= \left[\rho\sqrt{1 - \sigma} - \Delta T_2\left(\frac{c_1}{\sqrt{\alpha}} + 2c_2\sqrt{\gamma}\right)\right]\sqrt{V(t_2')}
	\end{align*}
	
	Integrating $\dot V$, we obtain 
	\begin{align*}
		V(t_2') - V(t_2) 
		&= \int_{t_2'}^{t_2} -\dot{V}(\tau)d\tau\\
		&\geq \int_{t_2'}^{t_2}\left(\|e(\tau)\|^2 + \frac{2\beta}{\gamma}\left\|\tilde{\theta}(\tau) - \tilde{\vartheta}(\tau)\right\|^2\right)d \tau\\
		&\geq \Delta T_2 \left[\rho\sqrt{1 - \sigma} - \Delta T_2\left(\frac{c_1}{\sqrt{\alpha}} + 2c_2\sqrt{\gamma}\right)\right]^2 V(t_2')
	\end{align*}
	Therefore, $V(t_2) \leq \eta V(t_2')$ and uniform asymptotic stability holds.
	
	Case 2: $\frac{1}{\gamma}\left\|\tilde{\theta}(t_2') - \tilde{\vartheta}(t_2')\right\|^2 \geq (1 - \nu)V(t_2')$. For any $t \geq t_2'$, following the process in case 1, we can show that
	\begin{align*}
		\left\|\tilde{\theta}(t_2') - \tilde{\vartheta}(t_2')\right\| - \left\|\tilde{\theta}(t) - \tilde{\vartheta}(t)\right\|
		\leq (t - t_2')\left(\beta\sqrt{\gamma} + \frac{c_2\gamma\rho}{\sqrt{\alpha}}\right)\sqrt{V(t_2')}
	\end{align*}
	If we let $t_2 = t_2' + \Delta T_2$, then
	\begin{align*}
		\left\|\tilde{\theta}(t) - \tilde{\vartheta}(t)\right\|
		&\geq \left\|\tilde{\theta}(t_2') - \tilde{\vartheta}(t_2')\right\| - (t_2 - t_2')\left(\beta\sqrt{\gamma} + \frac{c_2\gamma\rho}{\sqrt{\alpha}}\right)\sqrt{V(t_2')}\\
		&\geq \left[\sqrt{\gamma(1 - \nu)} - \Delta T_2\left(\beta\sqrt{\gamma} + \frac{c_2\gamma\rho}{\sqrt{\alpha}}\right)\right]\sqrt{V(t_2')}
	\end{align*}
	Integrating $\dot V$, we have
	\begin{align*}
		V(t_2') - V(t_2)
		\geq \frac{2\beta\Delta T_2}{\gamma}\left[\sqrt{\gamma(1 - \nu)} - \Delta T_2\left(\beta\sqrt{\gamma} + \frac{c_2\gamma\rho}{\sqrt{\alpha}}\right)\right]^2V(t_2')
	\end{align*}
	Therefore, $V(t_2)\leq\eta V(t_2')$ which proves the theorem.

\subsection{Proof of Theorem \ref{t:HAC-RL2}}
When $d(t)\neq 0$, with $f_1$ as in \eqref{eq:mod}, we obtain additional terms in $\dot V$, in the right-hand-side of \eqref{idealVdot} of the form (for ease of exposition, $\gamma$ is set to unity)
$$2e^TPB\Lambda d -2 Tr\left(\left(2\widetilde\Xi-\widetilde\theta\right)^T(\Lambda+\Lambda^T)\left(2\Xi-\Theta\right)\right)$$
where $\widetilde\Xi=\Xi-\Theta$ and $\widetilde\Theta=\widehat\Theta-\Theta$. Together with these terms, $\dot V$  can be shown to be negative outside an ellipsoid $Z^T\overline A Z=D_0$ where $Z=[\norm{e}, \norm{\widetilde \Xi -\widetilde\Theta},\norm{\widetilde\Xi}]^T$, $\overline A$ is a positive definite matrix in $\RR^{3\times 3}$ and $D_0$ is a constant that depends on $\bar d_0$ and $||\Theta||$ (the constant unknown parameter).

\subsection{Proof of Theorem \ref{t:MSAC-RL}}
As in Theorem 4, we consider a candidate
\begin{equation}
	\label{eq:lyap2}
	\begin{split}
		V &= \frac{1}{\gamma}\tr\left[(\Xi - \Theta)^T {\Lambda}^T (\Xi - \Theta)\right] + \frac{1}{\gamma}\tr\left[(\ov\Theta - \Xi)^T {\Lambda}^T (\ov\Theta - \Xi)\right] + e_u^T Pe_u
	\end{split}
\end{equation}
Using \eqref{eq:new-error} and \eqref{eq:msac_eqs}, we obtain that 
$\dot{V} \leq -e_u^T e_u$ since $Q_H \geq 2I$. 
This in turn allows us to conclude that $e_u,\Xi,\ov\Theta\in\linfty$.
For case (i), since all parameters are bounded, together with magnitude saturation, we have that the input $u$ is bounded. For this case, it implies that the state $x$ is bounded. Therefore $e_u\in\linfty$. Proceeding in a similar fashion to the proof of Theorem \ref{th:classic}, Barbalat's lemma is applied to find that $\lim_{t \to \infty} \|e_u(t)\| = 0$. It is easy to show then that 
$\|e(t)\| = O[\int_0^t\|\Delta u(\tau)\|d\tau]$.

For case (ii), as the structure of the error model in \eqref{eq:new-error} is identical to that considered in Theorem 1 in \cite{Sch05}, the same arguments can be used to establish boundedness of the state.

\subsection{Proof of Theorem~\ref{t:Contraction-HTAC}}
The first variation of the Riemannien energy \cite{do2013riemannian} is
\begin{align*}
\dot{E} &= \frac{\partial E}{\partial t} + \frac{\partial E}{\partial \hat{\nu}}\dot{\hat{\nu}} + 2\langle \gamma_s(s), \dot{\gamma}(s) \rangle \rvert_{s=0}^{s=1} - 2\int_0^1 \langle \frac{D}{ds} \gamma_s, \dot{\gamma}\rangle ds
\end{align*}
where $\frac{D(\cdot)}{ds}$ is the covariant derivative. Since $\gamma(s)$ is the geodesic, $\frac{D \gamma_s}{ds} = 0$.

Under Assumption~\ref{assum:extended_match}, Lemma 2 in \cite{lopez2020adaptive} implies
\begin{align*}
\frac{\partial E}{\partial \hat{\nu}}\dot{\hat{\nu}}+ 2\gamma_s(1)^T M(X, \hat{\nu}, t) B(X, t)U_{\nu} = 0
\end{align*}
Therefore, using \eqref{equ:HTAC-prob3} and \eqref{equ:u_c}, we have
\begin{align*}
\dot{E} &= \frac{\partial E}{\partial t} + \frac{\partial E}{\partial \hat{\nu}}\dot{\hat{\nu}} + 2\gamma_s(1)^T M(X, \hat{\nu}, t) \dot{X}\\
&\quad - 2\gamma_s(0)^T M(X_r, \hat{\nu}, t)\dot{X}_r\\
&= \frac{\partial E}{\partial t} + \left[\frac{\partial E}{\partial \hat{\nu}}\dot{\hat{\nu}}+ 2\gamma_s(1)^T M(X, \hat{\nu}, t) B(X, t)U_{\nu} \right]\\
&\quad + 2\gamma_s(1)^T M(X, \hat{\nu}, t) B(X, t) (\Lambda\Psi - I) U_{\nu}\\ 
&\quad+ 2\gamma_s(1)^T M(X, \hat{\nu}, t) \big[f(X, t) -\rho(X, t)\nu\\
&\quad+ B(X, t)(\Lambda \Psi - I) (U_c + \phi(X, t)\hat{\theta}) \big] \\
&\quad +2\gamma_s(1)^T M(X, \hat{\nu}, t) B(X, t)(U_c + \phi(X, t)\Tilde{\theta})\\
&\quad- 2\gamma_s(0)^T M(X_r, \hat{\nu}, t)\dot{X}_r\\
&\leq -2\lambda E + 2\gamma_s(1)^T M(X, \hat{\nu}, t) \big[ \rho(X, t) \Tilde{\nu}\\
&\quad+ B(X, t)((\Lambda \Psi - I) U + \phi(X, t)\Tilde{\theta})\big]
\end{align*}
Consider a Lyapunov candidate
{
\begin{align*}
	V &= \frac{1}{\gamma_{\nu}}(\nu_a - \hat{\nu})^T(\nu_a - \hat{\nu}) + \frac{1}{\gamma_{\nu}}(\nu_a - \nu)^T(\nu_a - \nu)\\
	&\quad+ \frac{1}{\gamma_{\theta}}(\theta_a - \hat{\theta})^T(\theta_a - \hat{\theta})
	+ \frac{1}{\gamma_{\theta}}(\theta_a - \theta)^T(\theta_a - \theta)\\
	&\quad+ \frac{1}{\gamma_{\Psi}} \tr\left[(\Lambda \Psi_a - \Lambda \Psi)^T |\Lambda|^{-1}(\Lambda
	\Psi_a- \Lambda\Psi)\right]\\
	&\quad+ \frac{1}{\gamma_{\Psi}} \tr\left[(\Lambda \Psi_a - I)^T |\Lambda|^{-1}(\Lambda \Psi_a- I)\right] + \frac{1}{2} E(\gamma, \hat{\nu}, t)^2 
\end{align*}
}
Define $e = \gamma_s(1) M E$, and use short-hand notation $\rho = \rho(X, t), \phi = \phi(X, t)$, and $B = B(X, t)$. Since the Riemmanien energy is always non-negative, i.e. $E \geq 0$, we have
\begin{align*}
\frac{1}{2}\dot{V} 
&\leq -\lambda E^2 + e^T \rho (\nu_a - \nu) + e^T B\big[(\Lambda \Psi - I) U\\
&\quad+ \phi(\theta_a - \theta)\big] -(\nu_a - \hat{\nu})^T \rho^T e - \frac{\beta_{\nu}}{\gamma_{\nu}}(\nu_a - \hat{\nu})^T(\nu_a - \hat{\nu})\mathcal{N}_{\nu}\\
&\quad- (\nu_a - \nu)^T\rho^T e -(\theta_a - \hat{\theta})^T\phi^T B^T e - \frac{\beta_{\theta}}{\gamma_{\theta}}(\theta_a - \hat{\theta})^T(\theta_a - \hat{\theta})\mathcal{N}_{\theta}\\
&\quad - (\theta_a - \theta)^T\phi^T B^T e + \tr\left[ -u e^T B(\Lambda \Psi_a - I)  \right]\\
&\quad+ \tr\left[ -U e^T B\Lambda (\Psi_a - \Psi) - \frac{\beta_{\Psi}}{\gamma_{\Psi}} \mathcal{N}_{\Psi}^T(\Psi_a - \Psi)^T \Lambda(\Psi_a - \Psi) \right]
\end{align*}
which can be simplified as $\dot V \leq 2(V_1+V_2+V_3) $, where 
\begin{align*}
V_1 &= -\frac{\lambda}{3} E^2 + 2e^T\rho (\hat{\nu} - \nu_a) - \frac{\beta_{\nu}}{\gamma_{\nu}}(\nu_a - \hat{\nu})^T(\nu_a - \hat{\nu})\mathcal{N}_{\nu}\\
V_2 &= -\frac{\lambda}{3} E^2 + 2e^TB\phi(\hat{\theta} - \theta_a) - \frac{\beta_{\theta}}{\gamma_{\theta}}(\theta_a - \hat{\theta})^T(\theta_a - \hat{\theta})\mathcal{N}_{\theta}\\
V_3 &= -\frac{\lambda}{3} E^2 + 2e^TB\Lambda(\Psi - \Psi_a)U -\frac{\beta_{\Psi}}{\gamma_{\Psi}} \tr\left[(\Psi_a - \Psi)^T \Lambda (\Psi_a - \Psi)\mathcal{N}_{\Psi} \right]
\end{align*}
It follows from \eqref{equ:HTAC-reg} and \eqref{equ:HTAC-P2-Gains} that
\begin{align*}
V_1 &\leq -\frac{\beta_{\nu}}{\gamma_{\nu}}\|\nu_a - \hat{\nu}\|^2 - \lambda_1(E - 2\|\omega_{\nu}\|\|\nu_a - \hat{\nu}\|)^2\\
V_2 &\leq -\frac{\beta_{\theta}}{\gamma_{\theta}} \|\theta_a - \hat{\theta}\|^2 - \lambda_2(E - \|\omega_{\theta}\| \|\theta_a - \hat{\theta}\|)^2 \leq 0
\end{align*}

%
Using similar manipulations as in the proof of Theorem 4, 
\begin{align*}
V_3 &\leq -\frac{\beta_{\Psi}}{\gamma_{\Psi}} \|(\Psi_a - \Psi)\Omega \|_F^2 - \lambda_3(E - \|\gamma_s(1)^T M B\| E \|(\Psi_a - \Psi)\Omega\|_F\|U\|)^2 \leq 0
\end{align*}

Thus we have $\dot{V} \leq 2(V_1 + V_2 + V_3)\leq 0$, which implies that $E, \hat{\nu}, \hat{\theta}, \Psi \in \mathcal{L}_{\infty}$, and that $E \in \mathcal{L}_2$. Since $\theta$ is a constant vector and $\Tilde{\theta}$ is bounded, $\hat{\theta}$ is bounded. Similarly, $\tilde{\nu}$ and $(\Lambda \Psi - I)$ are also bounded. By Assumption~\ref{assum:parameter_CCM}, $M(x, \hat{\nu}, t)$ is bounded. Since geodesics have constant speed and $E = \int_0^1 \gamma_s(s)^T M(x, \hat{\nu}, t) \gamma_s(s) ds = \langle \gamma_s, \gamma_s \rangle$ is bounded, $\gamma_s$ is bounded. Therefore, $\dot{E}\in \mathcal{L}_{\infty}$. By Barbalat's Lemma, 
$x(t) \to x_d(t)$ as $t \to \infty$. 

\bibliographystyle{IEEEtran}
\bibliography{IEEEabrv, References.bib}

\small
\textbf{Anuradha Annaswamy} is Founder and Director of the Active-Adaptive Control Laboratory (AACL) in the Department of Mechanical Engineering at the Massachusetts Institute of Technology (MIT), Cambridge, MA, USA.  Her research interests span adaptive control theory and its applications to aerospace, automotive, propulsion, and energy systems. 
She has received best paper awards (Axelby; CSM), Distinguished Member, and Distinguished Lecturer awards from the IEEE Control Systems Society (CSS) and a Presidential Young Investigator award from NSF. She is the author of a graduate textbook on adaptive control, co-editor of two vision documents on smart grids and the Impact of Control Technology report, and a coauthor of the National Academy of Sciences Committee report on the Future of Electric Power in the United States in 2021. She is a Fellow of IEEE and IFAC. She served as the President of CSS in 2020.

\noindent\textbf{Anubhav Guha} is a Masters student in the AACL at MIT. His research interests and focus include adaptive control, reinforcement and machine learning, and robotics.

\noindent\textbf{Yingnan Cui} is a PhD student in the AACL in the Department of Mechanical Engineering at MIT. His research interests include adaptive control and optimization methods for machine learning.

\noindent\textbf{Sunbochen Tang} is a PhD student in the AACL at MIT. His research interests include adaptive control, optimization methods, and machine learning.

\noindent\textbf{Peter Fisher} is a Masters student in the AACL at MIT. His research interests include adaptive control, optimization methods, and statistical learning.

\noindent\textbf{Joseph Gaudio} received his PhD degree in Mechanical Engineering from MIT in 2020. He is currently a Staff Engineer at Aurora Flight Sciences, a Boeing Company. His research interests are in adaptive control and machine learning.
\end{document}